\documentclass[10pt,twocolumn,letterpaper]{article}

\usepackage{multicol}
\usepackage{wacv}
\usepackage{times}
\usepackage{epsfig}
\usepackage{graphicx}
\usepackage{amsmath}
\usepackage{amssymb}
\usepackage{booktabs}
\usepackage{tabularx}
\usepackage{makecell}
\usepackage{lineno}
\usepackage{multirow}

\makeatletter
\@namedef{ver@everyshi.sty}{}
\makeatother
\usepackage{tikz}

\usepackage[accsupp]{axessibility}  

\def\wacvPaperID{449} 

\wacvfinalcopy 

\ifwacvfinal
\usepackage[breaklinks=true,bookmarks=false]{hyperref}
\else
\usepackage[pagebackref=true,breaklinks=true,colorlinks,bookmarks=false]{hyperref}
\fi

\usepackage{amsmath}
\usepackage{amssymb}
\usepackage{mathtools}
\usepackage{amsthm}

\usepackage[capitalize,noabbrev]{cleveref}
\usepackage{mathtools}
\usepackage{algorithm}
\usepackage{algpseudocode}
\usepackage{float}
\usepackage{color}
\usepackage{colortbl}
\usepackage{rotating}
\usepackage{times}
\usepackage{epsfig}
\usepackage{enumitem}
\usepackage{makecell}
\usepackage{lineno}
\usepackage{tabularx}
\usepackage{xcolor}
\usepackage{multirow}
\usepackage{graphicx}
\usepackage{hhline}
\usepackage{textcomp,booktabs}
\usepackage{caption}
\usepackage{stmaryrd}
\usepackage[mathscr]{eucal}
\usepackage{lineno}
\usepackage[export]{adjustbox}

\def\0{{\bf 0}}
\def\1{{\bf 1}}

\theoremstyle{plain}

\theoremstyle{definition}

\theoremstyle{remark}

\usepackage[textsize=tiny]{todonotes}

\usepackage[symbol]{footmisc}

\definecolor{red}{rgb}{0.95,0.4,0.4}
\definecolor{purered}{rgb}{1,0,0}
\definecolor{blue}{rgb}{0.4,0.4,0.95}
\definecolor{darkblue}{rgb}{0,0,0.8}
\definecolor{darkred}{rgb}{1,0,0}
\definecolor{darkgreen}{rgb}{0,0.5,0}
\definecolor{grey}{rgb}{0.6,0.6,0.6}
\definecolor{col1}{RGB}{232, 161, 148}
\definecolor{col2}{RGB}{148, 187, 232}
\definecolor{lightgrey}{rgb}{0.85,0.85,0.85}
\definecolor{lightlightgrey}{rgb}{0.9,0.9,0.9}
\definecolor{verylightBG}{rgb}{0.9,0.99,0.99}
\definecolor{darkgreen}{rgb}{0.3, 0.75, 0.3}

\begin{document}

\title{Creating a Forensic Database of Shoeprints from Online Shoe-Tread Photos}

\author{
Samia Shafique\textsuperscript{1} \quad\quad
Bailey Kong\textsuperscript{2} \quad\quad
Shu Kong\textsuperscript{3}\thanks{Authors share senior authorship.} \quad\quad
Charless Fowlkes\textsuperscript{1$\ast$}    
\\ 
\textsuperscript{1}University of California, Irvine \quad\quad  
\textsuperscript{2}Ronin Institute  \quad\quad  
\textsuperscript{3}Texas A\&M University 
\\
{\small {\tt sshafiqu@uci.edu} \quad
{\tt bailey.kong@ronininstitute.org} \quad 
{\tt fowlkes@ics.uci.edu} \quad 
{\tt shu@tamu.edu}} 
\\
{\small \url{https://github.com/Samia067/ShoeRinsics}}
\vspace{-3.5mm}
}

\maketitle
\thispagestyle{empty}

\begin{abstract}
   Shoe-tread impressions are one of the most common types of evidence left at crime scenes. However, the utility of such evidence is limited by the lack of databases of footwear prints that cover the large and growing number of distinct shoe models. Moreover, the database is preferred to contain the 3D shape, or depth, of shoe-tread photos so as to allow for extracting shoeprints to match a query (crime-scene) print.
   We propose to address this gap by leveraging shoe-tread photos collected by online retailers. The core challenge is to predict depth maps for these photos. As they do not have ground-truth 3D shapes allowing for training depth predictors, we exploit synthetic data that does. We develop a method, termed {\bf ShoeRinsics}, that learns to predict depth from fully supervised synthetic data and unsupervised retail image data. In particular, we find domain adaptation and intrinsic image decomposition techniques effectively mitigate the synthetic-real domain gap and yield significantly better depth predictions. 
   To validate our method, we introduce 2 validation sets consisting of shoe-tread image and print pairs and define a benchmarking protocol to quantify the quality of predicted depth. 
   On this benchmark, ShoeRinsics outperforms existing methods of depth prediction and synthetic-to-real domain adaptation. 
\end{abstract}

\section{Introduction}

\begin{figure}[t]
\center
  \includegraphics[trim={0 4.6cm 0 2.2cm},clip,width=\linewidth]{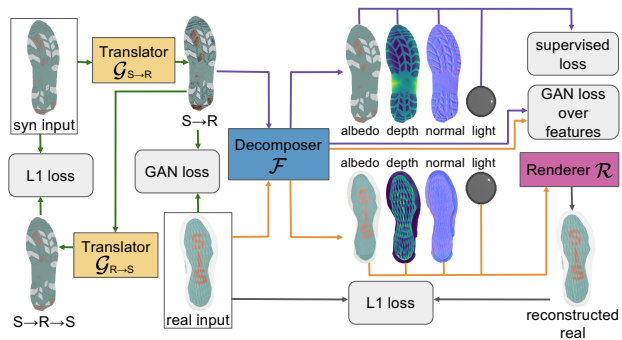}
  \vspace{-6mm}
  \caption{\small
  Predicting depth for shoe-tread images (collected by online retailers) is the core challenge in constructing a shoeprint database for forensic use. 
  We develop a method termed \emph{ShoeRinsics} to learn depth predictors.
   The flowchart depicts how we train \emph{ShoeRinsics} using annotated synthetic and un-annotated real images (Sec.~\ref{sec:data-prep}).
   We use domain adaptation (via image translators ${\mathcal G}_{S\rightarrow R}$ and ${\mathcal G}_{R\rightarrow S}$) and intrinsic image decomposition (via decomposer $\mathcal F$ and renderer $\mathcal R$) techniques to mitigate synthetic-real domain gaps (Sec.~\ref{sec:methodology}).
   Our method achieves significantly better depth prediction on real shoe-tread images than the prior art (Sec.~\ref{sec:exp}).
}
\vspace{-4mm}
\label{fig:architecture}
\end{figure}


Studying the evidence left at a crime scene aids investigators in identifying criminals. 
Shoeprints have a greater chance of being present at crime scenes \cite{bodziak2017footwear}, although they may have fewer uniquely identifying characteristics than other biometric samples (such as blood or hair). Thus, studying shoeprints can provide valuable clues to help investigators narrow down suspects of a crime.

Forensic analysis of shoeprints can provide clues on the {\em class characteristics} and the {\em acquired characteristics} of the suspect's shoe. The former involves the type of shoe (e.g., the brand, model, and size); the latter consists of the individual traits of a particular shoe that appear over time as it is worn (e.g., holes, cuts, and scratches). We are interested in aiding the study of class characteristics of shoeprints. 

{\bf Status quo.} Traditionally, investigating class characteristics of shoeprints involve matching the prints against a {\em manually curated} database of impressions of various shoe models \cite{bowen2007forensic}. 
The research community has shown significant interest in automating this matching process
\cite{bouridane2000application,de2005automated,gueham2007automatic,gueham2008automatic_class,gueham2008automatic_recog,algarni2008novel,jing2009novel,kortylewski2016probabilistic,kong2019cross,wang2017manifold,zhang2017adapting}. 
However, in practice, the success of such work depends on the quality of the database to which the shoeprint evidence is compared.
Yet, maintaining and regularly updating such a database to include all shoe models is tedious, costly, and requires significant human effort. 
Shoeprint matching methods are decidedly less useful if the database does not include the type of shoe the criminal wore!
Partly because of this, shoeprint evidence is vastly underutilized in the USA~\cite{smith2009}.

{\bf Motivation.} 
To address the need for such a comprehensive database, we propose to leverage imagery of shoe-treads collected by online retailers. High-resolution tread photos of various shoe products are readily available, and shopping websites are updated frequently ($>$1000 new products appear each month based on our analysis on some websites). Fig.~\ref{fig:syn_real_shoes} (b) shows examples of such shoe-tread images. 
Developing a method to predict the 3D shape from a shoe-tread image would directly address the need for a comprehensive, up-to-date database of tread patterns. 
\emph{We formulate this problem as depth prediction for shoe-treads; thresholding the depth map of a given shoe can generate/simulate shoeprints sufficient for matching query prints.}

{\bf Technical Insights.} 
To learn depth predictors from single shoe-tread images, we would ideally utilize supervised training examples of aligned shoe-tread images and their corresponding depth maps. However, since such ground-truth data is simply unavailable, we develop an alternative strategy. 
We create a synthetic dataset of rendered shoe-tread images and corresponding ground-truth depth, albedo, normal, and lighting. This data can train a predictor in a fully supervised fashion. However, the resulting model performs sub-optimally on real-world images due to the domain gap between synthetic and real imagery.
To address this, we introduce three additional techniques to close the synthetic-real domain gap by incorporating methods of domain adaptation~\cite{CycleGAN2017} and intrinsic image decomposition~\cite{nips2017}
(see Fig.~\ref{fig:architecture}). 
First, we train a translator that translates synthetic shoe-treads to realistic images, which better match the distribution of the real shoe-treads. 
Second, we use an adversarial loss to enforce that features of real and translated synthetic images are indistinguishable.
Third, we use a re-rendering loss that adopts a synthetically trained renderer to reconstruct the real shoe-tread images using their predicted depth and other intrinsic components. 
We find these three techniques in combination help close the domain gap and yield significantly better depth prediction.
{\bf Contributions.} We make three major contributions.
\begin{itemize}[noitemsep,  topsep=0pt]
    \item 
    Motivated to create a database of shoeprints
    for forensic use, we introduce the task of depth prediction for real shoe-tread photos collected by online retailers.
    \item We develop a benchmarking protocol, with which we evaluate existing methods of depth prediction using domain adaptation for this task.
    \item We develop a method called \emph{ShoeRinsics} that incorporates intrinsic image decomposition and domain adaptation techniques, outperforming prior art for this task. 
\end{itemize}

\section{Related Work}

{\bf Shoeprint Analysis.}
Automatic shoeprint matching has been studied widely in the past two decades \cite{rida2019forensic}. Existing works focus on generating good features from shoeprints and using them to assign a class label (shoe type) from a database of lab footwear impressions. 
To study global features (i.e., considering the whole shoe), \cite{kortylewski2016probabilistic} introduces a probabilistic compositional active basis model, \cite{kong2019cross} explores multi-channel normalized cross-correlation to match multi-channel deep features, and \cite{wang2017manifold} employs a manifold ranking method, and \cite{zhang2017adapting} uses VGG16 as a feature extractor. 
On the other hand, \cite{ma2019shoe} studies a multi-part weighted CNN, \cite{alizadeh2017automatic} introduces a block sparse representation technique, and \cite{almaadeed2015partial} applies multiple point-of-interest detectors and SIFT descriptors to study the local features of shoeprints (i.e., keypoints \cite{krig2016interest}).
Our work differs from the previous work as it focuses on creating a database of prints rather than developing methods for shoeprint matching.   \emph{Creating such as database is a prerequisite for algorithmic explorations for shoe-matching.}

{\bf Monocular Depth Prediction} has been studied extensively since early works \cite{hoiem2005automatic,saxena20083,saxena2005learning}. Previous methods invent features representations~\cite{baig2016coupled,ranftl2016dense,furukawa2017depth},  deep network architectures~\cite{alp2017densereg,li2017two,roy2016monocular,kendall2017uncertainties,laina2016deeper,xie2016deep3d}, and training losses~\cite{fu2018deep,shin20193d,yin2019enforcing}.  \cite{kuznietsov2017semi,garg2016unsupervised,li2018megadepth} explore self-supervised learning in a stereo setup while \cite{ranftl2020towards,zhao2020domain} experiment with training on large datasets. Depth estimation has been further improved by considering the camera pose \cite{zhao2021camera}.
Our work differs from the above as it aims for depth prediction on real images by learning over un-annotated real images and synthetic images (and their ground-truth intrinsics: depth, albedo, normal and light).

{\bf Intrinsic Image Decomposition.}
Another line of work aims to explain image appearance in terms of some intrinsic components, including albedo, normals, and lighting. However, predicting intrinsic images is difficult, if not impossible. Our approach is related to \cite{nips2017}, which learns for intrinsic image decomposition and uses a differentiable renderer to leverage un-annotated images with a reconstruction loss. 
\cite{sfsnetSengupta18,phy_face_relighting2020,multi_channel_portrait_relight_2020} focus on face images and explore a similar reconstruction loop~\cite{sfsnetSengupta18}, non-diffuse lighting models~\cite{phy_face_relighting2020}, and multiple reflectance channels~\cite{multi_channel_portrait_relight_2020}.
\cite{wu2021derender} works on rotationally symmetric objects with only object silhouettes as supervision. 
\cite{neuralSengupta19,YuSelfRelight20,glosh,li2020inverse,zhu2021cvpr} study decomposition on entire scenes.
\cite{alhaija2021intrinsic} learns photo-realistic rendering of synthetic data and intrinsic decomposition of real images using unpaired data as input via an adversarial loss. 
In contrast, our work utilizes intrinsic decomposition techniques to help learn depth prediction by leveraging annotated synthetic and un-annotated real data via domain adaptation.

{\bf Domain Adaptation.}
Training solely on synthetic data can cause models to perform poorly on real data. Adversarial domain adaptation has proved promising for bridging such domain gaps. 
One way to approach this is to use domain-invariant features to map between the domains.
\cite{long2013transfer}  proposes to reduce the Maximum Mean Discrepancy to learn domain-invariant features. \cite{tzeng2014deep} builds on this idea and further improves domain adaptation performance in classification tasks. \cite{adda,tsai2018learning,udab,sun2016deep} learn domain adaptation by aligning source and target features. 
Another direction of work uses image-to-image translation \cite{CycleGAN2017}  to stylize source images as target images. \cite{cycada, zheng2018t2net} use the stylized source images to learn from target images using source labels while performing alignment both at the image and feature level.
We use domain adaptation for depth estimation but take this approach further by reasoning about the intrinsic components of unlabeled real data.

\begin{table}[t] 
\scriptsize
\caption{\small 
Overview of our datasets for training and testing, along with their shoe categories  and counts. 
It is worth noting that real-val contains formal and used shoes, which are not present in training (i.e., the real-train set). We include these novel shoe types to analyze the generalizability of different methods. See details in Sec.~\ref{sec:data-prep} and visual examples in  Fig.~\ref{fig:syn_real_shoes}.
}
\vspace{-2mm}
\setlength\tabcolsep{0pt} 
\begin{tabular*}{\linewidth}{>{}p{1.3cm} @{\extracolsep{\fill}}*{5}{c}}
\toprule
Dataset & \multicolumn{3}{c}{Shoe Category} & Total & Annotation \\ 
\cmidrule{2-4}
& New-Athletic    &   Formal  &  Used    \\
\midrule
syn-train   &   88,408 &   0    &   0  & 88,408 &  depth,  albedo, normal, light \\
real-train  &   3,543 &   0    &   0   & 3,543 & none \\
real-val    &   22 &   6   & 8  & 36 & print \\
real-FID-val   &   41 &   0    &   0   & 41 & print \\
\bottomrule
\end{tabular*}
\vspace{-4mm}
\label{table:dataset_statistics}
\end{table}

\section{Problem Setup and Evaluation Protocol}
Our motivation is to create a database of shoeprints for forensic use. 
\emph{The specific task is to predict depth maps for shoe-tread images collected by  online retailers.} Below, we formulate the problem and introduce an evaluation protocol to benchmark methods.

\subsection{Problem Setup}
Online shoe-tread photos do not have ground-truth depth. Thus, we cannot directly train a depth predictor on them.
Instead, we propose to create a dataset of synthetic shoe-tread images for which we have a complete set of annotations, including depth, albedo, normal, and lighting (details in Section~\ref{ssec:syn-train}).
Therefore, {\bf the problem is to predict depth for \emph{real} shoe-treads by learning a depth predictor on synthetic shoe-treads (with annotations) and real shoe-treads (without annotations)}.
This requires (1) learning a depth predictor by exploiting synthetic data that has annotations of depth and other intrinsic components, (2) addressing the synthetic-real domain gap.

\subsection{Evaluation Protocol}
\label{sec:metric}

Recall that the created database, containing predicted depth maps and shoe-tread images, and will serve for forensic use --  an investigator will query a shoeprint collected at a crime scene by matching it with depth maps within this database. Therefore, we evaluate the quality of predicted depth maps w.r.t shoeprint matching. 

To this end, we introduce two validation sets 
that contains paired ``ground-truth'' shoeprints and shoe-tread photos (details in Section~\ref{ssec:real-val}).
For a given shoe-tread, a trained model predicts its depth and the metric measures the degree of match between the ground-truth shoeprint and the predicted depth.
We develop a metric based on Intersection-over-Union (IoU).
Specifically, we generate a set of shoeprints using adaptive thresholding (with a range of hyperparameters) for the predicted depth, and compute the IoU between the ground-truth print to each of these generated shoeprints. The metric returns the highest IoU.
We further average the IoUs over all the validation data as mean IoU (mIoU) to benchmark methods.
Refer to the supplement for further details.


\begin{figure}[t]
\center
\includegraphics[trim={0 6.3cm 0 3.5cm},clip,width=\linewidth]{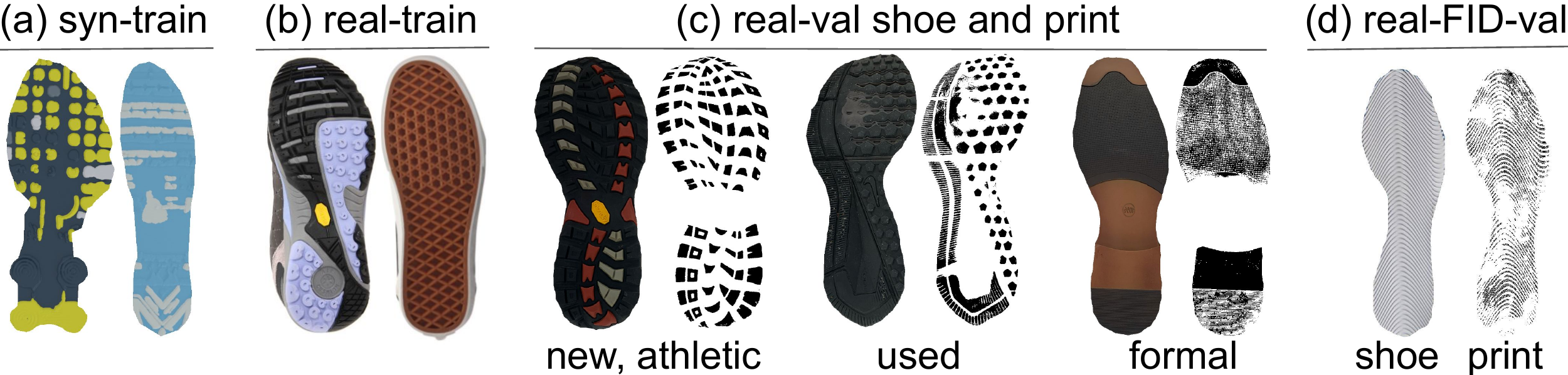}
\vspace{-6mm}
\caption{\small 
Shoe tread examples from (a) syn-train, (b) real-train, (c) real-val, and (d) real-FID-val. 
Clearly, a domain gap exists between (a) syn-train and (b) real-train, demonstrating the need to close the synthetic-real domain gap. 
Moreover, to study the generalizability, we evaluate on 2 datasets (c) and (d) and purposely hold out the formal and used shoe-treads which are not used for training but for validation (c).
}
\vspace{-4mm}
\label{fig:syn_real_shoes}
\end{figure}

\section{Data Preparation}
\label{sec:data-prep}

During training, we have two data sources: a synthetic dataset (\emph{syn-train}) that has annotations, and a dataset of un-annotated real shoe-treads (\emph{real-train}).
To study models' generalizability, we test our model on two validation sets  (\emph{real-val} and \emph{real-FID-val}).
Each of these datasets contain shoe-tread photos with aligned ground-truth shoeprints,
which enable quantitative evaluation.
Note that to analyze the models' robustness to novel shoe types, we constrain our training sets to contain only brand-new athletic shoes while letting real-val also include formal and used (worn) shoes.
Fig.~\ref{fig:syn_real_shoes} displays example shoe-treads and Table \ref{table:dataset_statistics} summarizes the four datasets.
Below, we elaborate on the creation of the synthetic training set (syn-train), the real training set (real-train), and validation sets (real-val and real-FID-val).


\subsection{Synthetic Data for Training}
\label{ssec:syn-train}

\begin{figure}[t]
\center
\includegraphics[trim={0 13.4cm 2.9cm 2.2cm},clip,width=\linewidth]{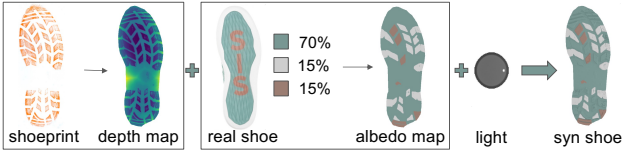}
\vspace{-6mm}
\caption{\small
Generation of synthetic data.
We scale off-the-shelf shoeprints to generate ``pseudo'' depth maps. 
We sample a color distribution from a real-shoe example to create an albedo map. The depth map and albedo map are combined with a lighting environment to render a synthetic image. The lighting environment is demonstrated by visualizing a shiny sphere in place of the shoe. In this example, directional light comes from a point on the right.
}
\vspace{-4mm}
\label{fig:syntheticdataset}
\end{figure}

Our synthetic dataset (syn-train) containing synthetic shoe-tread images and their intrinsic annotations (depth, albedo, normal, and lighting). 
We synthesize a shoe-tread image with a given depth map, an albedo map, and a lighting environment (outlined in Fig.~\ref{fig:syntheticdataset}).
We pass these to a physically-based rendering engine~\cite{Mitsuba} to generate the synthetic image.
The final syn-train set contains 88,408 shoe-treads with paired ground-truth intrinsic images. 

{\bf Depth Map.} 
We use an existing dataset~\cite{isreal-shoe} to generate plausible synthetic depth maps to create syn-train data.
For each of 387 shoeprints, we synthesize 10-15 different depth maps.
Because the shoeprints have noise that affects synthetic data generation, we first apply a Gaussian blur to filter the noise. 
We then scale the blurred print image to create a ``pseudo'' depth map.
To generate more diverse depth maps we add random high-frequency textures.
Lastly, we make tread shapes more realistic by adding a priori features, such as slanted bevels on the tread elements and global curvature of the shoe-tread (details in supplement).

{\bf Albedo Map.} The color palette for each rendered shoe comes from the color distribution of a real shoe-tread photograph. Shoes tend to have only a handful of different colors across the entire tread. We identify the primary colors on real shoe-treads using the mean-shift algorithm~\cite{fukunaga1975estimation}. 
Albedo maps for the rendered shoes are composed of these colors. First, we use depth maps to identify shoe-tread elements and segment out areas of the shoe that can have different colors. 
Then we assign colors to those segments from the color palette of a real shoe in the percentages in which they are present. 
Fig.~\ref{fig:syntheticdataset} shows one example.

{\bf Light environment.} Online retail stores use specialized diffuse lighting rigs to capture shoe photos. We create a similar lighting environment for our rendered images. Shoes are photographed with bright diffuse white light from all directions and some optional directional light. 
We use a total of 17 different light configurations. One light configuration is simply diffuse light coming from all directions. Eight light configurations consist of single light bulbs shining from eight directions around the shoe in addition to the diffuse white light. The remaining eight are similar but contains two light bulbs at \(120^{\circ}\) to each other. 
The supplement has further details.

\subsection{Online Shoe Treads for Training and Prediction}
\label{ssec:real-val}

Online retailers \cite{Sixpm,Zappos} adopt photos of shoes for advertisement, which include shoe-tread images. Real-train (3,543), cf. Table~\ref{table:dataset_statistics}, consists of such shoe-tread images and masks computed by a simple network to segment out the shoe-treads. This dataset does not contain any ground-truth and consists only of new, athletic shoes.

\begin{figure}[t]
{\small \hspace{3mm}shoe-tread \hspace{3mm} pseudo albedo \hspace{8mm} shoe-tread \hspace{3mm} pseudo albedo}\\
\vspace{-6mm}
\center
  \includegraphics[width=\linewidth]{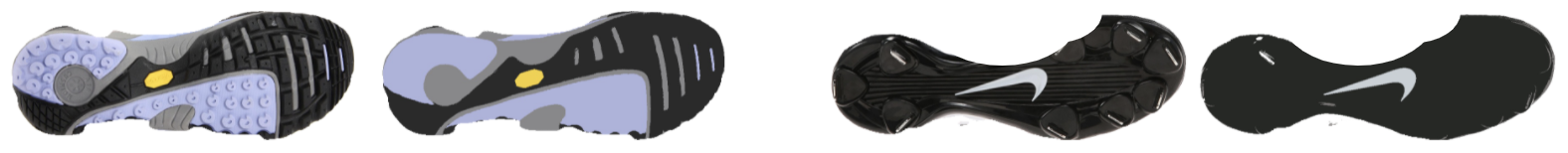}
\vspace{-7mm}
   \caption{\small
   Generating pseudo albedo maps from shoe-tread images. We show two pairs.
   We run the mean-shift algorithm~\cite{fukunaga1975estimation} on a shoe-tread image to group RGB pixels, resulting in the corresponding pseudo albedo map.
   We use the pseudo albedo maps as supervision signals to train the decomposer (cf. Fig.~\ref{fig:architecture}).
   }
\vspace{-4mm}
\label{fig:pseudo_albedo}
\end{figure}

\subsection{Lab Data for Validation}

\begin{figure}[!t]
\center
  \includegraphics[width=\linewidth,trim={0 13.6cm 34cm 0},clip]{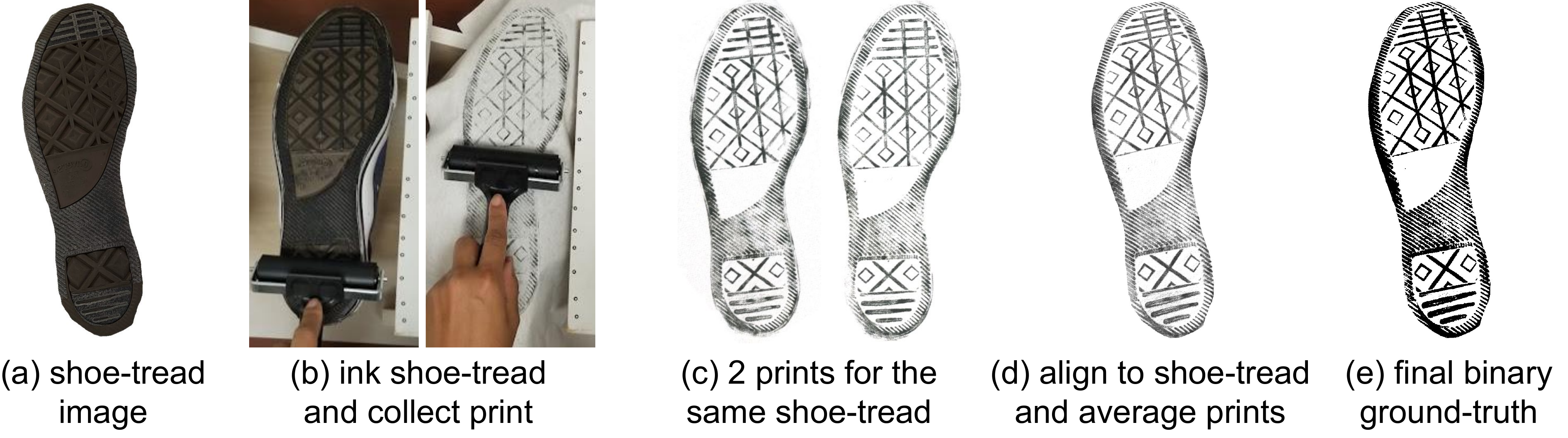}
\vspace{-8mm}
\caption{\small
We collect a validation set of ground-truth shoeprints from shoes in a lab environment. (a) shows an example shoe. (b) It is painted with a thin layer of relief ink, and a paper sheet is pressed evenly onto the shoe-tread using a roller. (c) We repeat this to get 2-3 different prints. (d) We align these prints to the shoe-tread using thin-plate spline~\cite{duchon1977splines} and (e) threshold their average to obtain the final ground-truth shoeprint, which has better coverage. 
}
\vspace{-4mm}
\label{fig:print_collection}
\end{figure}

{\bf Real-val}.
To quantitatively benchmark methods, we collect paired shoe-tread images and ground-truth prints in a lab environment. Fig.~\ref{fig:print_collection} summarizes the procedure. 
We photograph shoes by placing them inside a light box with a ring light on top. We collect prints from those shoes by painting the treads with a thin layer of relief ink and pressing absorbent white papers onto the shoe-treads. This method of collecting shoeprints is called the {\em block printing technique} and is one of several techniques used in the forensics community to collect reference footwear impressions~\cite{bodziak2017footwear}.
To improve print quality, we collect 2-3 prints for each shoe and average them after alignment to the shoe-tread.
We use thin-plate splines~\cite{duchon1977splines} with a smoothness parameter of 0.5 for alignment. 
We threshold the average print as the final ground-truth shoeprint.
Real-val contains 22 new-athletic shoes, 6 new formal shoes, and 8 used athletic shoes. The formal and used shoes are not present during training and thus serve as novel examples in evaluation.

{\bf Real-FID-val}.
We introduce the second validation set consisting of shoeprints from the FID300 dataset~\cite{fid300} and shoe-tread images separately downloaded from online retailers (i.e., these images are disjoint from those in the real-train set).
We find matched FID300 prints (used as the ground-truth) and the downloaded shoe-tread images, and align them manually. 
Real-FID-val contains 41 new, athletic shoe-tread images with corresponding ground-truth shoeprints and masks to segment out the shoe-treads.

\section{Methodology}
\label{sec:methodology}

We now introduce our \emph{ShoeRensics}, a pipeline that trains a depth predictor for real images $I_R$ by incorporating unsupervised adversarial domain adaptation and intrinsic image decomposition techniques. 
Given synthetic images \(I_{S}\) with their corresponding ground-truth intrinsics (albedo \(X_{S}^a\), depth \(X_{S}^d\), normal \(X_{S}^n\), and light \(X_{S}^l\)) and unlabeled real images \(I_{R}\), our goal is to train a model to predict depth $d_{R}$ for real images $I_R$. 
Fig.~\ref{fig:architecture} overviews our training pipeline.
The main components of our pipeline are a translator \({\mathcal G}_{S\rightarrow R}\) to stylize synthetic images as real images, a decomposer $\mathcal F$ for intrinsic image decomposition, and a renderer $\mathcal R$ to reconstruct the input images from their intrinsic components.


{\bf Synthetic-only Training.}
We train a decomposer $\mathcal F$ and the renderer $\mathcal R$ in a supervised manner on syn-train.
For an input image, the decomposer predicts depth $\hat X_S^d$, albedo $\hat X_S^a$, normal $\hat X_S^n$, and light ${\hat X}_S^l$.
The renderer $\mathcal R$ learns to reconstruct the input image from these predicted intrinsic components.
To train the decomposer $\mathcal F$, we use an $\mathcal{L}_{1}$ loss to learn for depth, albedo, and normal prediction and a cross-entropy loss $\mathcal{L}_{CE}$ to learn for light (treating light prediction as a  $K$-way classification problem given the limited light sources).
We minimize the overall loss below:
\vspace{-2mm}
{\small
\begin{align}
&\mathcal{L}_{sup} =  \lambda_l \mathcal{L}_{CE}({\hat X}_S^l, X_{S}^{l}) + \sum_{\kappa \in \{d, a, n\}} \lambda_{\kappa} \mathcal{L}_{1}({\hat X}_{S}^{\kappa}, X_{S}^{\kappa})
\label{eq:syn-loss}.
\end{align}
} %
where $\lambda$'s are hyperparamters controlling loss terms for the intrinsic components.
To learn the renderer $\mathcal R$, we simply minimize the $\mathcal{L}_{1}$ loss between the original and rendered images, i.e., $\mathcal{L}_{1}(I_S, {\mathcal R}(X_S^d, X_S^a, X_S^n, X_S^l))$.
Note that depth prediction is our main focus, and we find learning with  decomposer and renderer significantly helps depth learning (cf. Fig.~\ref{fig:architecture}, Table~\ref{table:iou}).
A model trained on synthetic data only does not work effectively well on real data due to the notorious synthetic-real domain gap. We address this issue using the techniques below.

{\bf Mitigating domain gap by image translation.}
Previous work \cite{cycada,CycleGAN2017} addresses the domain gap between image sources by translating images from one domain to the other. We adopt a similar approach and translate our synthetic images to realistic ones by training a translator \({\mathcal G}_{S\rightarrow R}\).
We train another ${\mathcal G}_{R\rightarrow S}$ that translates real images to synthetic style. Discriminators 
$\mathcal D_{R}(I)$ and $\mathcal D_S(I)$ 
are learned simultaneously to discriminate translated images and used for training translators. This is known as the adversarial domain adaptation~\cite{cycada}.
We further translate the translated synthetic/real images back to the original domain and use a cycle loss between the resulting and the initial images 
to ensure that structure and content are preserved during translation. 
The following losses train the translators~\cite{cycada,CycleGAN2017}:
\vspace{-1mm}
{\small
\begin{align}
&\mathcal{L}_{GAN}^{S\rightarrow R}(I_R, I_S) = \log \mathcal D_{R}(I_{R})  + \log(1 - \mathcal D_{R}(\mathcal G_{S\rightarrow R}(I_{S}))) \nonumber\\
&\mathcal{L}_{GAN}^{R\rightarrow S}(I_S, I_R) = \log \mathcal D_{S}(I_{S})  + \log(1 - \mathcal D_{S}(\mathcal G_{R\rightarrow S}(I_{R}))) \nonumber\\
&\mathcal{L}_{tran} = \mathcal{L}_{GAN}^{S\rightarrow R}(I_{R}, I_{S})  + \mathcal{L}_{GAN}^{R\rightarrow S}(I_{S}, I_{R}) \nonumber\\ 
&\mathcal{L}_{cyc}  =  \mathcal{L}_{1}(\mathcal G_{R\rightarrow S}(\mathcal G_{S\rightarrow R} (I_{S})), I_{S}) + \nonumber\\
& \quad \quad \quad \ \ 
\mathcal{L}_{1}(\mathcal G_{S\rightarrow R}(\mathcal G_{R\rightarrow S} (I_{R})), I_{R}) 
\label{eq:translator-losses}
\end{align}
}%
With $\mathcal G_{S\rightarrow R}(I_S)$, we translate syn-train images and keep their corresponding ground-truth intrinsics unchanged.
We use such translated data to finetune the renderer $\mathcal R$.

{\bf  Mitigating domain gap by image reconstruction.}
We additionally use an image reconstruction loss to address the domain gap~\cite{nips2017}.
We reconstruct a real image from its decomposed intrinsic components using the trained renderer $\mathcal R$, which we freeze after finetuning on translated synthetic data.
We use $\mathcal R$ to regularize the training of the decomposer $\mathcal D$ on real images.
Denoting reconstructed real image as  $\hat I_{R}\coloneqq \mathcal R(X_{R}^d, X_{R}^a, X_R^n, X_R^l)$, we minimize the difference between the original image $I_{R}$ and its reconstruction $\hat I_R$ using an $\mathcal{L}_{1}$ loss, i.e.,  $\mathcal{L}_{1}(\hat I_R, I_{R})$.

\begin{figure*}[th]
\center
\includegraphics[width=0.80\linewidth,trim={0 9.1cm 0.7cm 0},clip]{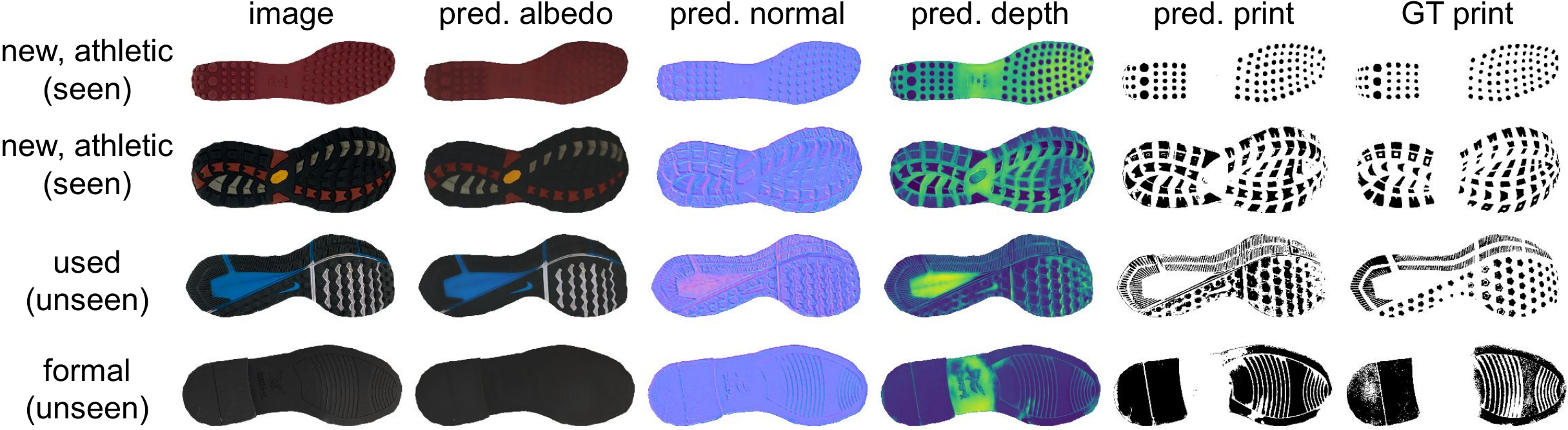}
\vspace{-2mm}
\caption{\small
   On images of the real-val set, we visualize \emph{ShoeRinsics}'s predictions including depth thresholding which generates predicted prints.  
   Our method \emph{ShoeRinsics} produces visually appealing intrinsic decompositions (depth, albedo, and normal).
   Importantly, on novel shoe-tread displayed in the last bottom rows, \emph{ShoeRinsics} produces very good depth and shoeprints by comparing against the ground-truth shoeprints.
   To display the predicted prints, we threshold the predicted depth to best match the ground-truth print (Sec~\ref{sec:metric}).
   }
\vspace{-4mm}
\label{fig:visual_for_val_shoe_ours}
\end{figure*}

{\bf  Mitigating domain gap by feature alignment.}
We further adopt the feature alignment technique to mitigate the domain gap~\cite{CycleGAN2017}. Specifically,
we learn an adversarial discriminator \(\mathcal D_{feat}\) to discriminate \emph{features} extracted by the decomposer for the real images and the translated synthetic images. We use this as a loss in training the decomposer and update the discriminator $\mathcal D_{feat}$ while training of the decomposer. This encourages the decomposer to extract features on real data that are indistinguishable from synthetic data, thus helping mitigate the domain gap.


{\bf Exploiting Pseudo Albedo.}
Shoe-treads, like many other man-made objects such as cars and other toys, tend to have piece-wise constant albedo. 
Building on this observation, we create pseudo albedo for the real data by grouping pixels with the mean-shift algorithm~\cite{fukunaga1975estimation}. 
Fig.~\ref{fig:pseudo_albedo} shows an example pseudo albedo on two real shoes.
As pseudo albedo is not ideal as ground-truth, we use it to learn an albedo predictor through the the decomposer. We find  this produces better albedo maps than the pseudo ground-truth (see analysis in the supplement).
To learn albedo prediction, we minimize the $\mathcal{L}_{1}$ loss, i.e., $\mathcal{L}_{1}(\hat{X}_{R}^a, \text{MS}(I_{R}))$, where MS is the mean-shift clustering algorithm.

{\bf Stage-wise Training} is common in training multiple modules, particularly with GAN discriminators.
Our training paradigm contains four stages.
First, we train the decomposer \(\mathcal F\) and renderer \(\mathcal R\) on syn-train. 
Second, we train the image translators and discriminators \(G_{S\rightarrow R}\), \(\mathcal G_{R\rightarrow S}\), \(\mathcal D_{R}\), and \(\mathcal D_{S}\) with Eq.~\ref{eq:translator-losses}.
Third, we finetune $\mathcal R$ using the translated synthetic images by \(\mathcal G_{S\rightarrow R}\). 
Finally, we freeze \(\mathcal R\) and \(\mathcal G_{S\rightarrow R}\) and finetune $\mathcal F$ on translated synthetic images and real images using losses described above.

\section{Experiments}
\label{sec:exp}

We validate our \emph{ShoeRinsics} and compare it against prior methods of depth prediction on our benchmark.
We start with implementation details, followed by a visual comparison and quantitative evaluation, and conduct an ablation study and  analysis of why \emph{ShoeRinsics} outperforms the prior art. 

\subsection{Implementation}

{\bf Training specifics}.
Instead of using high-resolution images (405x765) from the training set, we crop patches (128x128) to train the models. 
We find this yields better performance, as shown in the ablation study (Sec.~\ref{ssec:ablation}).
For a fair comparison, we train all models with patches for the same number of optimization steps.
During training, we sample patches from random positions. 
We use Adam optimizer and set the learning rate as 1e-3 and 1e-4 for training the initial models (e.g.,  $\mathcal F$ and $\mathcal R$) and finetuning them, respectively. 
We set the batch size as 8 throughout our experiments.
Recall that we train our model in stages (Sec.~\ref{sec:methodology}).
We train for 20M iterations in the first two stages and 100K iterations in the last two stages.

{\bf Architectures}.
Our decomposer $\mathcal F$  and renderer $\mathcal R$ have a classic encoder-decoder structure as used in \cite{nips2017}. We modify the light prediction decoder to be a 17-way classifier (given that our synthetic data has only 17 lighting configurations).
We also add residual connections between layers to predict full-resolution maps for intrinsic components (depth, albedo, and normal). 
Our translators and discriminators (\({\mathcal G}_{S\rightarrow R}, {\mathcal G}_{R\rightarrow S}, {\mathcal D}_{R}, {\mathcal D}_{S}\), and \({\mathcal D}_{feat}\)) have the same structure as used in \cite{cycada}. 
The \({\mathcal D}_{feat}\) is a convolutional network that uses a kernel size 3 to process the albedo, depth, and normal features. It further takes as input the features of the lighting prediction branch. That said, ${\mathcal D}_{feat}$ learns to discriminate features of all the intrinsic components.

{\bf Hyperparameter setting}.
We denote the combined hyperparameters  as $\hat\lambda = (\lambda_a, \lambda_d, \lambda_n, \lambda_l)$ in Eq.~\ref{eq:syn-loss}.
The decomposer $\mathcal F$ is trained with \(\hat\lambda = (1, 1, 1, 0.1)\) in the first stage, and finetuned with \(\hat\lambda = (1, 2, 1, 0.1)\) in the final stage.
When finetuning, we set the weight to  3 for the reconstruction loss, 2 for the pseudo albedo loss, and 1 for the feature alignment.
We set the hyperparameters via validation.



{\bf Test-time augmentation}.
During testing, we consider test-time augmentation~\cite{david2020intrinsic,he2016deep}.
For each image, we produce 23 variants: 3 flips (horizontal, vertical, and vertical+horizontal), 4 rotations (angles $+5^\circ$, $+10^\circ$, $-5^\circ$, and $-10^\circ$), 4 scalings (scale factor 0.5, 0.8, 1.5, and 1.8), and $12$ flip+rotation versions (three flips times four rotations). For each variant, we predict the depth and then transform back to the original coordinate frame.
We average all the 24 depth maps as the final prediction.

\begin{figure}[t]
\centering
  \includegraphics[width=0.93\linewidth,trim={0 0 55cm 0},clip]{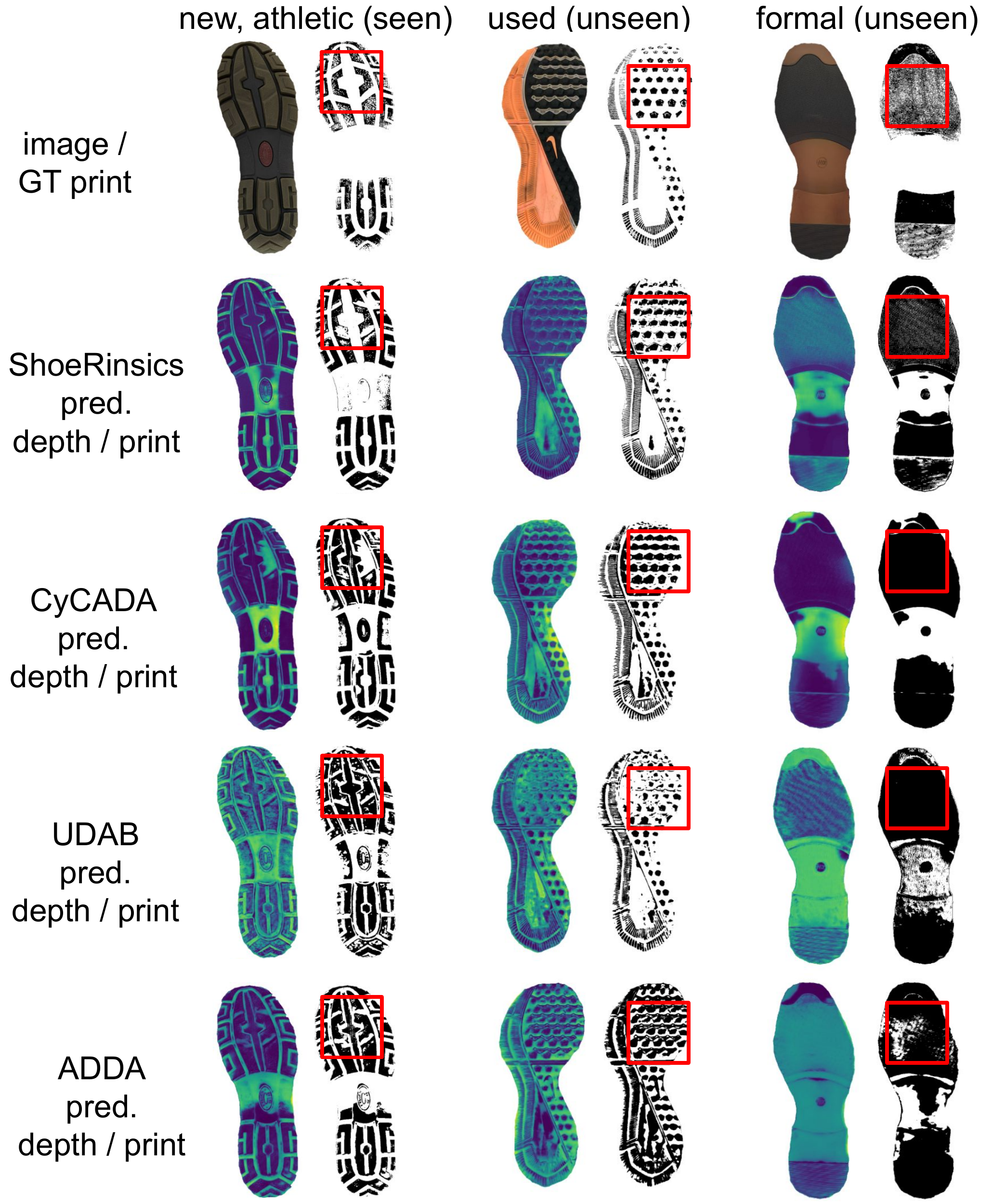}
\vspace{-4mm}
\caption{\small
   Comparison with the state-of-the-art methods of domain adaptation tailored to depth prediction on our real-val benchmark. 
   Our \emph{ShoeRinsics} performs better than others for both seen and unseen shoe categories as highlighted by the red boxes.
}
\vspace{-4mm}
\label{fig:visual_for_val_shoe_comparison}
\end{figure}

\begin{figure}[t]
\vspace{-0.5mm}
\center
\includegraphics[width=0.93\linewidth,trim={0 0 56.5cm 0},clip]{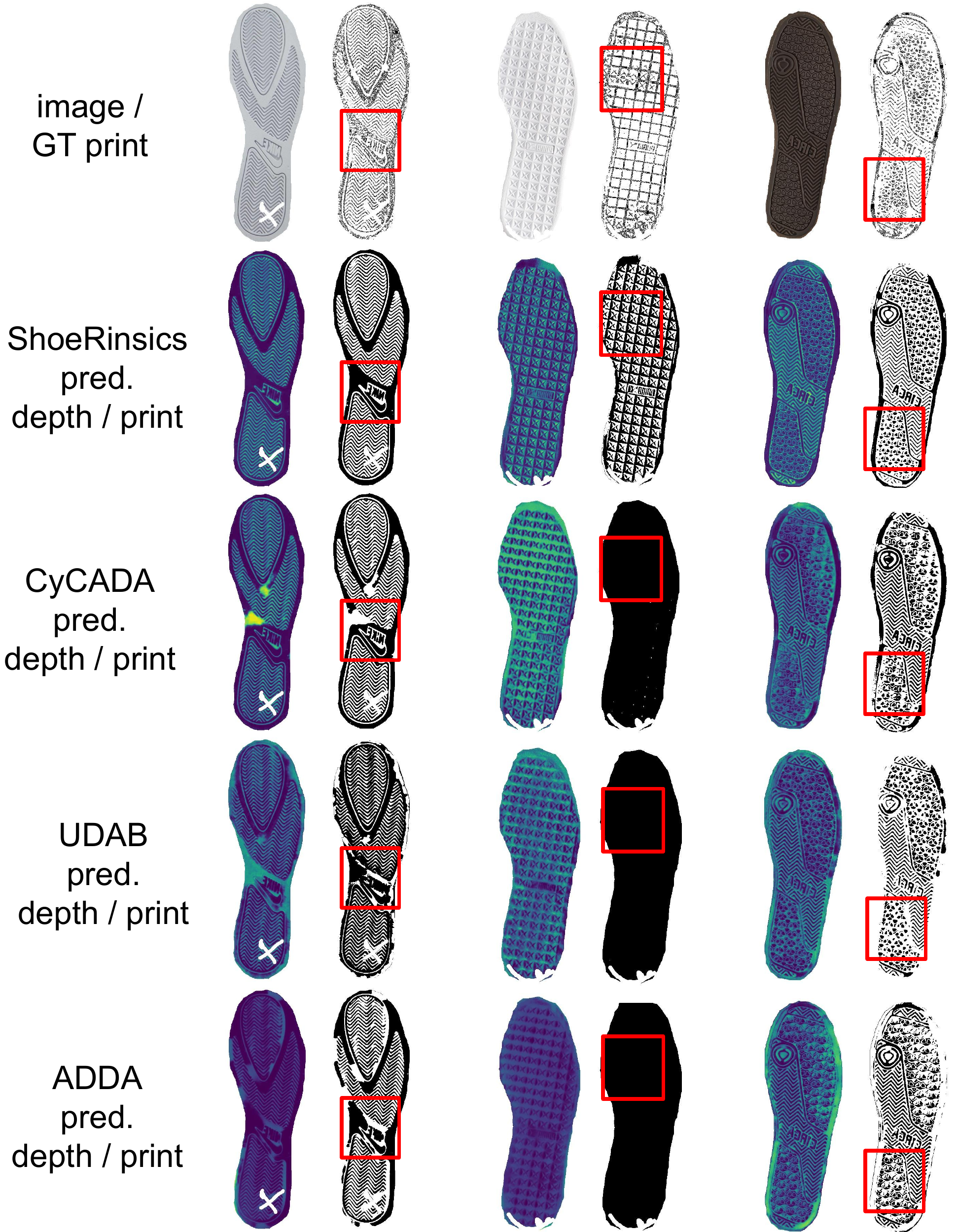}
\vspace{-4mm}
\caption{\small
Comparison with the state-of-the-art methods for depth prediction and domain adaptation~\cite{udab,cycada,adda} on real-FID-val. 
Clearly, our \emph{ShoeRinsics} produces shoeprints which are visually closer to the ground-truth than previous methods. 
}
\vspace{-4mm}
\label{fig:visual_for_real_FID_val}
\end{figure}

\subsection{Qualitative Results of ShoeRinsics}

We visualize predictions on the real-val images by our method \emph{ShoeRinsics} in Fig.~\ref{fig:visual_for_val_shoe_ours}. 
\emph{ShoeRinsics} predicts good depth maps, the thresholding of which generates shoeprints that match the ground-truth prints. 
As a byproduct, our method also makes visually appealing predictions on other intrinsic components. 
We compare our predictions with those made by other methods on real-val (Fig.~\ref{fig:visual_for_val_shoe_comparison}) and real-FID-val (Fig.~\ref{fig:visual_for_real_FID_val}).
Clearly, our \emph{ShoeRinsics} produces more reasonable visuals (depth and shoeprints) than the compared methods. 
The supplement has further visualizations.

\subsection{State-of-the-art Comparison}

\emph{ShoeRinsics} outperforms prior methods in most of the validation examples (details in the supplement). 
Table~\ref{table:iou} and \ref{table:iou_real_FID_val} list comparisons as analyzed below.

{
\setlength{\tabcolsep}{0.26em}
\begin{table} [t]
\centering
\caption{\small
Benchmarking on real-val. We use IoU as the metric (in \%), and break down the analysis for different shoe categories (\emph{new-athletic} shoes seen during training, and \emph{formal} and \emph{used} shoes unseen in training).
We compute mean IoU (mIoU) over all validation examples.
Training on only synthetic data yields poor performance, whereas our \emph{ShoeRinsics} performs the best on both seen and unseen categories. This clearly demonstrates the benefit of combining synthetic-to-real domain adaptation with intrinsic decomposition. 
The ablation study (bottom panel) shows that each individual component (discriminator, translator, and renderer, cf. Fig.~\ref{fig:architecture}) helps improve shoeprint prediction. Lastly, from our syn-only ablation, decomposing to all intrinsic components performs better than training a depth predictor for shoeprint prediction, further demonstrating that incorporating intrinsic decomposition helps close synthetic-to-real domain gaps.
Exploiting test-time augmentation boosts performance from {\em mIoU}=46.8 to 49.0.
}
\vspace{-3mm}
{\small
\begin{tabular}{lcccc}
\toprule                          
\multirow{2}{*}{\em Method}            & \em New-Athletic & \em Formal   & \em Used & \multirow{2}{*}{\em mIoU}\\
    & \em (seen) & \em (unseen)   & (unseen) &   \\
\midrule
RIN \cite{nips2017}     & 30.0  & 39.7  & 24.4  & 30.4 \\
ADDA \cite{adda}        & 46.5  & 41.4 & 27.2  & 41.4 \\
UDAB \cite{udab}        & 46.0  & 40.4  & 29.6  & 41.4  \\
CyCADA \cite{cycada}    & 48.8  & 43.9  & 34.5  & 44.8 \\  
\midrule
syn-only, depth only   & 41.3  & 41.2  & 28.4  & 38.4 \\
syn-only, all intrinsics    & 41.8  & 41.5  & 27.1  & 38.5 \\
\textbf{ShoeRinsics}          &  50.5     & 47.8    & 35.8 & 46.8 \\

\quad w/o discriminator        & 48.2  & 39.9  & 33.6  & 43.6 \\
\quad w/o translator          & 49.0  & 42.8  & 31.4  & 44.0 \\
\quad w/o renderer             & 49.0  & 46.4  & 34.7  & 45.4 \\
\textbf{ShoeRinsics} w/ aug        & \textbf{52.4}  & \textbf{52.9}  & \textbf{36.9} & \textbf{49.0} \\
\bottomrule
\end{tabular}
}
\vspace{-1mm}
\label{table:iou}
\end{table}
}

\setlength{\tabcolsep}{0.33em}
\begin{table} [t]
\centering
\caption{\small
Benchmarking on real-FID-val. We report mean IoU (mIoU) over validation examples.
\emph{ShoeRinsics} outperforms previous methods and improves further with test-time augmentation. 
}
\vspace{-3mm}
{\footnotesize
\begin{tabular*}{\linewidth}{lcccccc}
\toprule                          
 & RIN & ADDA & UDAB & CyCADA  & \textbf{ShoeRinsics} & \textbf{ShoeRinsics} \\
 & \cite{nips2017} & \cite{adda} & \cite{udab} & \cite{cycada} & &  w/ aug \\
 \midrule
 mIoU & 26.0 & 27.2 & 29.0 & 31.2 & 31.6 & \textbf{32.0}\\
\bottomrule
\end{tabular*}
}
\vspace{-4mm}
\label{table:iou_real_FID_val}
\end{table}

{\bf Comparison with intrinsic image decomposition}.
We compare our \emph{ShoeRinsics} and RIN~\cite{nips2017}, which learns for intrinsic image decomposition.
As RIN~\cite{nips2017} emphasizes normal prediction to represent shapes, we use the standard Frankot-Chellappa algorithm~\cite{frankot1988method} to integrate the normals towards depth maps.
Compared to \cite{nips2017}, our \emph{ShoeRinsics} explicitly incorporates domain adaptation in the image and feature space. Doing so helps mitigate the synthetic-real domain gap.
As a result, \emph{ShoeRinsics} outperforms RIN on both real-val and real-FID-val (Table~\ref{table:iou} and \ref{table:iou_real_FID_val}). 
On real-val, it performs better than RIN  by 20.5\% mIoU on the (\emph{seen new-athletic}) shoes,
by 8.1\% mIoU on the \emph{formal unseen} shoes,  
by 11.4\% mIoU on \emph{used unseen} shoes. 
On real-FID-val, \emph{ShoeRinsics} improves IoU by 5.6\% mIoU over RIN.

{\bf Comparison with domain adaptation}.
Table~\ref{table:iou} and \ref{table:iou_real_FID_val} clearly show that our \emph{ShoeRinsics} consistently outperforms the compared domain adaptation methods (ADDA~\cite{adda}, UDAB~\cite{udab}, and CyCADA~\cite{cycada}) on both the real-val and real-FID-val datasets.
From ablation studies, as shown in the lower panel of Table~\ref{table:iou}, we see that using the renderer (cf. Fig.~\ref{fig:architecture}) and the decomposer (that learns to predict albedo, normal, and lighting as auxiliary supervisions) greatly improves the performance.
Qualitative comparison on real-val in Fig.~\ref{fig:visual_for_val_shoe_comparison} and real-FID-val in Fig.~\ref{fig:visual_for_real_FID_val} show that depth maps and the corresponding prints predicted by our \emph{ShoeRinsics} have richer textures and better-aligned patterns to the RGB input.
When exploiting test-time augmentation (cf. \emph{ShoeRinsics} w/ test-time aug), we boost the performance  from mIoU = 46.8\% to 49.0\% on real-val and from mIoU=31.6\% to 32.0\% on real-FID-val.

\begin{figure}[t]
\centering
\includegraphics[width=0.8\linewidth,trim={0 12.7cm 39cm 0},clip]{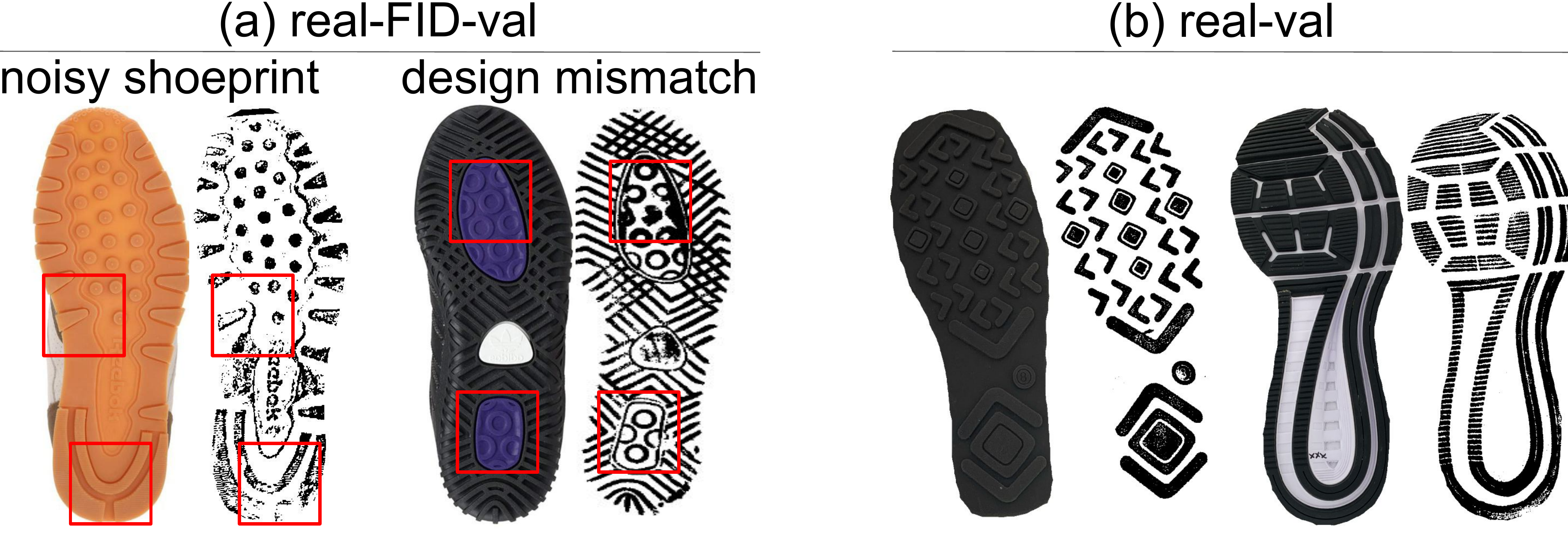}
\vspace{-3mm}
\caption{\small
Comparison between real-FID-val (a) and real-val (b). The shoeprints from real-FID-val are noisy and slightly misaligned with the corresponding shoe-treads. 
In contrast, shoeprints of real-val contain the entire contact surfaces and are well aligned with the corresponding shoe-tread images.
}
\vspace{-1mm}
\label{fig:FID_prints}
\end{figure}

{\bf Performance on real-val vs real-FID-val.}
All the methods show lower mIoU numbers on real-FID-val compared to real-val. 
This is owing to the noisy ground-truth prints of real-FID-val (see Fig.~\ref{fig:FID_prints}). 
Note that the FID prints are obtained by pressing gelatin lifters onto dusty shoe-treads followed by scanning the lifters~\cite{fid300}. 
This means that the shoeprints can be noisy as the contact surfaces do not leave a full print. 
In contrast, for real-val shoeprints, we minimize such noise and get more even coverage by averaging over multiple prints for the same shoe. 
Moreover, while real-val consists of image and print pairs of the exact same shoe, real-FID-val consists of prints from~\cite{fid300} with our manually discovered shoe-tread images, meaning that they might not be well aligned, as visually seen in Fig.~\ref{fig:FID_prints}.

\begin{figure}[!t]
\center
  \includegraphics[width=0.9\linewidth,trim={0 19cm 51.9cm 0},clip]{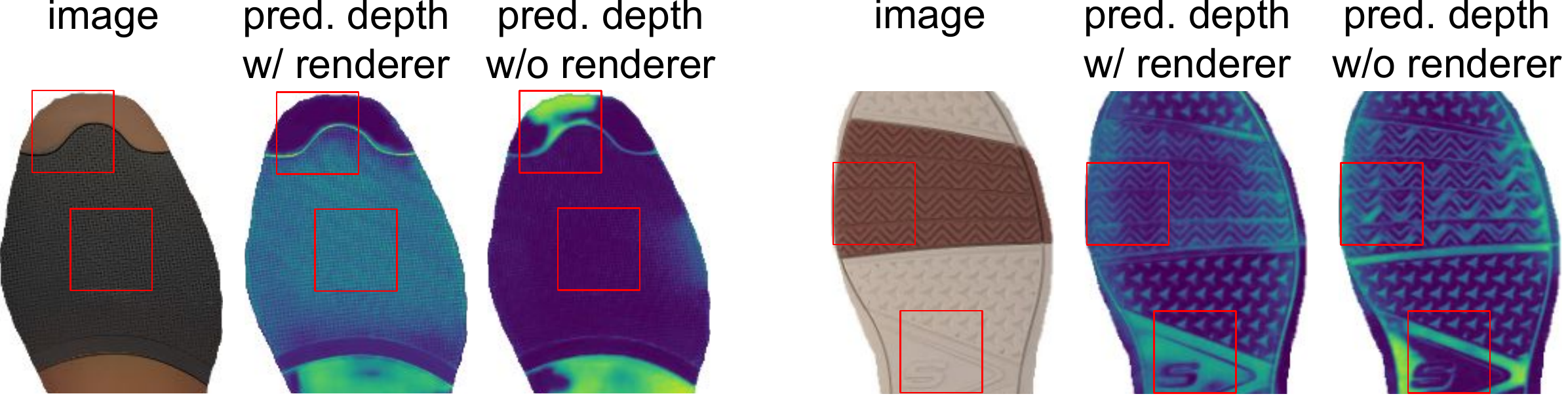}
\vspace{-4mm}
\caption{\small
Training \emph{ShoeRinsics} with the renderer (which allows using the reconstruction loss) produces visibly better depth than without.
Using the renderer encourages the decomposer to output depth maps that contain fine-grained details because it penalizes coarse predictions through the image reconstruction loss. That said, the renderer regularizes the learning for depth prediction by exploiting auxiliary supervisions from other intrinsic components (albedo, normal, and lighting).
}
\vspace{-4mm}
\label{fig:vis_renderer4domaingap}
\end{figure}

\subsection{Ablation Study}
\label{ssec:ablation}
We conduct an ablation study (cf. Table~\ref{table:iou} bottom panel)  on the modules in \emph{ShoeRinsics}, including feature alignment (by learning discriminator ${\mathcal D}_{feat}$ in the feature space), translator $\mathcal G_{S\rightarrow R}$, and renderer $\mathcal R$. All three modules aim to mitigate synthetic-real domain gaps.
We also study whether predicting intrinsic components (albedo, normal, and lighting) helps depth prediction and whether patch-based learning is better than full-image learning.


{\bf Effect of feature alignment by discriminator ${\mathcal D}_{feat}$}.
{\em ShoeRinsics w/o discriminator} removes the feature discriminator \({\mathcal D}_{feat}\) but keeps all the other modules. 
It yields 43.6\% mIoU, 3.2\% mIoU lower than \emph{ShoeRinsics} (cf. Table \ref{table:iou}). 
This demonstrates the effectiveness of \({\mathcal D}_{feat}\) for mitigating domain gaps by aligning features.


{\bf Effect of image translator $\mathcal G_{S \rightarrow R}$}. 
{\em ShoeRinsics w/o translator} drops the translators but keeps other components, achieving 44.0\% mIoU, 2.8\% mIoU lower than \emph{ShoeRinsics} (cf. Table \ref{table:iou}). This shows the effectiveness of using translators to close the synthetic-real domain gap.

{\bf Effect of the reconstruction loss by the renderer $\mathcal R$}.
{\em ShoeRinsics w/o renderer} drops the renderer from \emph{ShoeRinsics}, leading to 45.4\% mIoU, 1.4\% mIoU lower than  \emph{ShoeRinsics} (cf. Table \ref{table:iou}). This validates the effectiveness of the renderer.
Fig.~\ref{fig:vis_renderer4domaingap} visualizes depth predictions with and without the renderer during training. Clearly, with the renderer, the predicted depth has better high-frequency textures. See the caption of Fig.~\ref{fig:vis_renderer4domaingap} for details.


{\bf All intrinsics vs depth only}.
Comparing ``syn-only, depth only'' and ``syn-only, all intrinsics'' in Table~\ref{table:iou}, we see that learning to predict all intrinsics performs slightly better (38.5\% vs. 38.4\%). Importantly, this allows using the renderer as the reconstruction loss to regularize the training on real images, yielding significantly better results in the final \emph{ShoeRinsics} (46.8\% mIoU).

{\bf Patches vs. full-resolution images}.
We compare the depth prediction performance by training the decomposer on patches versus full-resolution images of the synthetic data.
We find that the former (patch-based) achieves 38.5\% mIoU (cf. Table \ref{table:iou}) as opposed to 36.5\% mIoU for the latter (not shown in the table).
This demonstrates the benefit of depth learning on patches over whole images in this setup.

\subsection{Failure Cases}

\begin{figure}[t]
{\footnotesize \hspace{4mm} image \hspace{4.5mm} pred. albedo \hspace{1.5mm} pred. normal \hspace{2mm} pred. depth \hspace{3mm} pred. print}
\vspace{-3mm}
\center
  \includegraphics[width=\linewidth,trim={0 0 0 2.2cm},clip] {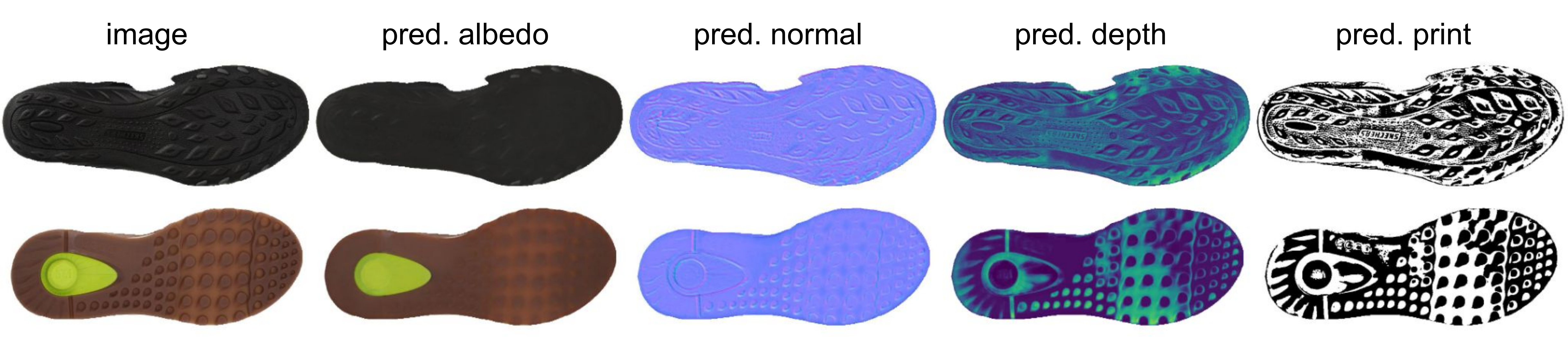}
\vspace{-7.5mm}
\caption{\small
   Failure cases. \emph{ShoeRinsics} performs poorly in the presence of complex materials (e.g., translucence). 
   }
\vspace{-4mm}
\label{fig:visual_fail}
\end{figure}

We analyze failure cases of \emph{ShoeRinsics} in Fig.~\ref{fig:visual_fail}.
We find that our method performs poorly on shoes with complex materials. One reason is that the syn-train data does not contain any complex materials.
Future work may explore richer synthetic datasets to improve performance.

\section{Conclusion}
Motivated by constructing a database of shoeprints  for forensic use, we introduce a problem of predicting depth for shoe-tread photos collected by online retailers.
Because these photos do not have ground-truth depth, we exploit synthetic images (containing shoe-treads and ground-truth intrinsics including depth, albedo, normal, and lighting).
We study domain adaptation and intrinsic image decomposition techniques and propose a method termed \emph{ShoeRinsics} to train  for depth prediction.
Our experiments demonstrate consistent improvements of \emph{ShoeRinsics} over previous methods on this task.
We expect future algorithmic explorations on this task from the perspective of domain adaptation, depth prediction, and intrinsic decomposition.

{\bf Acknowledgements}. This work was funded by the Center for Statistics and Applications in Forensic Evidence (CSAFE) through Cooperative Agreements, 70NANB15H176 and 70NANB20H019.



{\small
\bibliographystyle{ieee_fullname}
\bibliography{egbib}
}

\clearpage


\onecolumn
\begin{center}
{\Large\bf{Appendix}\\
}
\end{center}

\vspace{20pt}

\begin{multicols}{2}
\section*{Outline}

We propose to create a forensic database of shoeprints by leveraging shoe-tread imagery collected by online retailers. To do so, we strive to predict depth maps for these photos, thresholding whihc generates shoeprints used to match a query print (collected from a crime scene). We propose a novel method, \emph{ShoeRinsics}, to learn depth estimation from synthetic data (along with intrinsic components) and real data (with no annotations). \emph{SheoRinsics} incorporates synthetic-to-real domain adaptation and intrinsic image decomposition techniques to mitigate domain gaps. 
We validate our method with a defined evaluation protocol that measures the degree of match between predicted  depth and ground-truth shoeprints (collected in a lab environment).
Results convincingly demonstrate that \emph{ShoeRinsics} remarkably outperforms state-of-the-art methods for shoe-tread depth prediction.
In this supplementary document, we discuss the following topics:
\begin{itemize}
    \item Section~\ref{sec:supp_metric} details the process of matching a ground-truth shoeprint to predicted depth for evaluation.
    \item Section~\ref{sec:qualitaive_analysis} shows qualitative analysis of our \emph{ShoeRinsics}. We visualize of all predictions of \emph{ShoeRinsics} on real-FID-val images downloaded from the internet in Section~\ref{sec:results_real_test} and compare to RIN~\cite{nips2017} in Section~\ref{sec:supp_rin}. 
    \item Section~\ref{sec:quantitative_analysis_details} compares performance of \emph{ShoeRinsics} with related work for each individual image from real-val and real-FID-val.
    \item Section~\ref{sec:syn_data} provides details on our synthetic dataset generation. The process of depth map generation is described in Section~\ref{sec:syn_data_depth} and the light environemnts used are visualized in Section~\ref{sec:syn_data_light}.
    \item Section~\ref{sec:supp_gt_print} describes how we photograph shoe-treads and collect their prints to create a validation set (real-val) for quantitative evaluation.
    \item Section~\ref{sec:supp_implementation} details the architecture of each component of our \emph{ShoeRinsics}.
    \item Section~\ref{sec:supp_pseudo_albedo} discusses the pseudo albedo generated for real shoe-tread images and compares them to the albedo predicted by \emph{ShoeRinsics}.
\end{itemize}

{
\setlength{\textfloatsep}{0pt} 
\begin{algorithm}[H]
\small
\caption{Metric of Depth-Print Matching}
\begin{algorithmic}[1]
\State Input: predicted depth $\hat X^d_R$, ground-truth shoeprint $S^*$, and mask $m$
\State Initialize best-matching IoU $v_{max}=0$ 
\State determine per-pixel local depth, $d_l = \frac{\text{blurred depth}}{\text{blurred mask}} $ \\
\For{$s \in [0.1, 0.11, 0.12, \dots, 2]$}
\If{$\text{IoU}(\hat X^d_R < s d_l, S^*) > v_{max}$}
        \State $v_{max}=\text{IoU}(\hat X^d_R < s d_l, S^*)$, 
        \State set best-matching shoeprint $S_{best}=\hat X^d_R<sd_l$
\EndIf
\EndFor \\
\State set $p_{95}$ = value at the $95$th percentile in sorted $\hat X^d_R$
\For{$t_{nc} \in [0.1 p_{95}, 0.1 p_{95} + 0.01, 0.1 p_{95} + 0.02, \dots, p_{95}]$}
\State determine shoeprint $S_{t_{nc}} = S_{best}\: \text{AND}\: (\hat X^d_R < t_{nc})$
\If {$\text{IoU}(S_{t_{nc}}, S^*)>v_{max}$}
        \State $v_{max}=\text{IoU}(S_{t_{nc}}, S^*)$, \quad  $S_{best}=S_{t_{nc}}$
\EndIf
\EndFor \\
\State set $p_{05}$ = value at the $5$th percentile in sorted $\hat X^d_R$
\For{$t_{c} \in [p_{05}, p_{05} + 0.1, p_{05} + 0.2, \dots, 30p_{05}]$}
\State determine shoeprint $S_{t_{c}} = S_{best}\: \text{OR}\: (\hat X^d_R < t_{c})$
\If {$\text{IoU}(S_{t_{c}}, S^*)>v_{max}$}
        \State $v_{max}=\text{IoU}(S_{t_{c}}, S^*)$ 
\EndIf
\EndFor \\
\Return   best-matching IoU $v_{max}$
\end{algorithmic}
\label{alg:thresh-free-eval}
\end{algorithm}
}

\section{Depth-Print Matching in Evaluation}
\label{sec:supp_metric}

\begin{figure*}[!t]
\center
  \includegraphics[width=\linewidth,trim={0 21.3cm 56cm 0},clip]{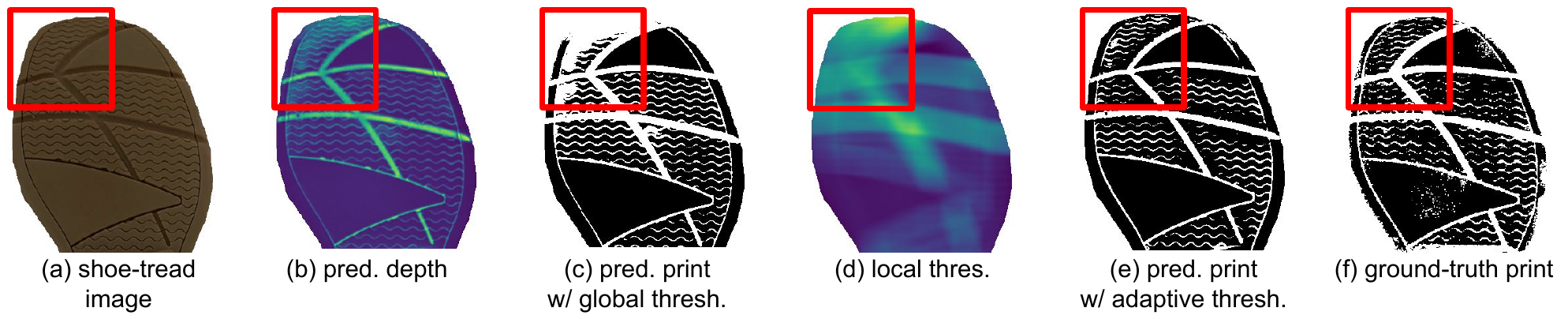}
\caption{\small
Effect of adaptive threshold. Given a shoe-tread image (a), we predict its depth (b). Notice that the front of the shoe is curved up slightly (highlighted by the red boxes). Thus, a global threshold fails to capture the print properly. Compare the print prediction using a global threshold (c) with the ground-truth print (f). A solution is to use an adaptive threshold instead. As such, we first blur the predicted depth to get the local mean depth $d_l$. Using $s d_l$ as the local threshold (d) where $s$ is an appropriate constant for scaling, we get predicted print (e) which is much closer to the ground-truth (f).
}
\label{fig:adaptive_threshold}
\end{figure*}

We get the shoeprint prediction by thresholding the predicted depth $\hat X^d_R$ of a shoe-tread image. However, different thresholds can produce different predicted shoeprints. So, we develop a threshold-free metric for evaluating how well our predicted depth matches a ground-truth shoeprint. Ideally, we want to threshold the depth prediction in a way that produces a shoeprint prediction that most closely matches the ground-truth shoeprint.
We summarize our method in Algorithm~\ref{alg:thresh-free-eval}.

\begin{figure}[H]
\center
  \includegraphics[width=\linewidth,trim={0 21.6cm 66.4cm 0},clip]{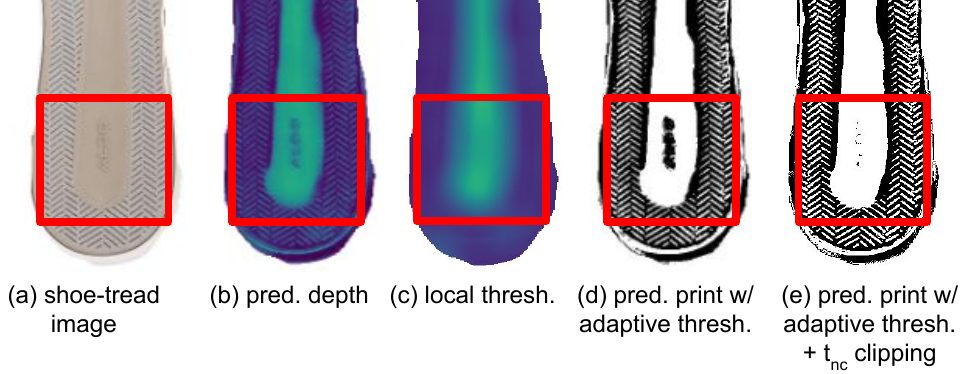}
\vspace{-4mm}
\caption{\small
Effect of non-contact threshold $t_{nc}$. A shoe-tread image (a), the corresponding depth prediction (b), the local threshold (c), and the predicted print prediction with adaptive threshold (d) is shown. 
We can see that using only the adaptive threshold can cause errors in large areas of non-contact surface as shown by the red boxes. 
Assume non-contact surfaces have high depth values. 
Although the logo is correctly predicted to have a high depth value, the local threshold also happens to be high in the region and causes adaptive thresholding to undesirably predict the logo to leave a print. 
To correct this, we find an appropriate non-contact threshold $t_{nc}$ for which regions where predicted depth is greater than $t_{nc}$ does not leave a print. The resulting print  is shown in (e). 
}
\label{fig:t_nc_effect}
\end{figure}

\begin{figure}[H]
\center
  \includegraphics[width=\linewidth,trim={0 19.7cm 61.4cm 0},clip]{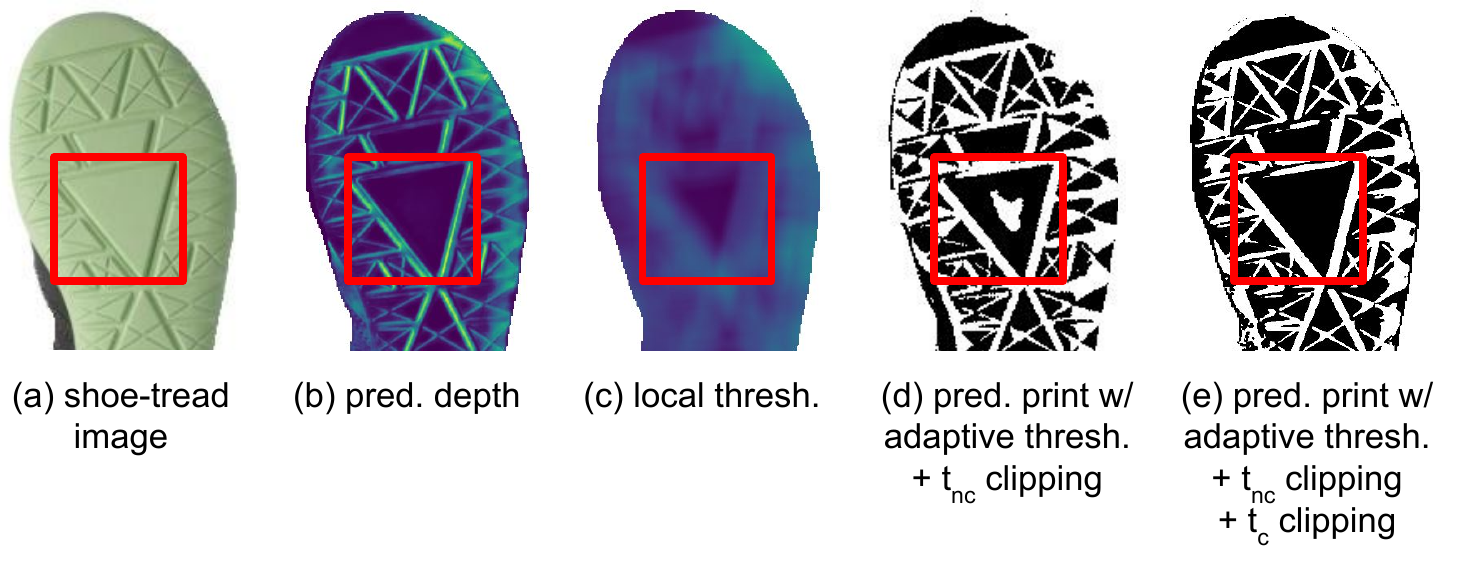}
\vspace{-6mm}
\caption{\small
Effect of contact threshold $t_c$. We visualize a sample shoe-tread image (a), the predicted depth (b), the local threshold (c), and the print prediction after using adaptive thresholing with $t_{nc}$ clipping (d). 
Parts of a large contact surface incorrectly leaves no print as shown by the red boxes because the local threshold is very low in the region (assuming contact areas have low depth value). To correct this, we determine an appropriate contact threshold $t_c$ such that regions where predicted depth is less than $t_c$ always leave a print. (e) demonstrates the final result. 
}
\vspace{-4mm}
\label{fig:t_c_effect}
\end{figure}

{\bf Global thresholding vs. adaptive thresholding.}
Using a global threshold for print prediction from depth prediction is troublesome since errors can creep into regions where the shoe-tread curves upwards (for example, in the front of the shoe). Fig.~\ref{fig:adaptive_threshold} illustrates this scenario with a sample shoe from real-val. 
In such cases, although the shoe is curved upwards, it still leaves a print when someone walks wearing those shoes. This is because the weight of the person flattens out the shoe. Also, the physical motions of walking causes the curved parts to come in contact with the ground. 
Assuming high depth values correspond to non-contact surfaces, a portion of the shoe that curves up would have high depth values and a global threshold might incorrectly indicate that region does not leave a print. We can solve this issue by using adaptive thresholding instead. 

{\bf Details of adaptive thresholding.}
To perform adaptive thresholding, we first determine a per-pixel local average $d_l$ for the predicted depth. 
This is achieved by blurring the predicted depth $\hat X^d_R$ with a large square kernel of size $45 \times 45$. For comparison, our shoe-tread image and predicted depth map resolution is $405 \times 765$.
We note that boundaries and invalid depth values outside the mask may cause artifacts in $d_l$. 
To negate this effect, we set $d_l = \frac{d_l}{m_l}$ where $m_l$ is the per-pixel local average for the mask computed in a similar manner.

Next, we set our best-matching shoeprint $S_{best} = \hat X^d_R < s d_l$ for some scalar multiplier $s$. Theoretically, we want to sweep over all possible values of $s$ and find the one which gives the  highest IoU between $\hat X^d_R < s d_l$ and the ground-truth print $S^*$. Practically, we sample values from range $[0.1, 2]$ at intervals of $0.01$. 

\begin{figure*}[t]
\centering
\includegraphics[width=\linewidth,trim={0 0 33.5cm 0 },clip]{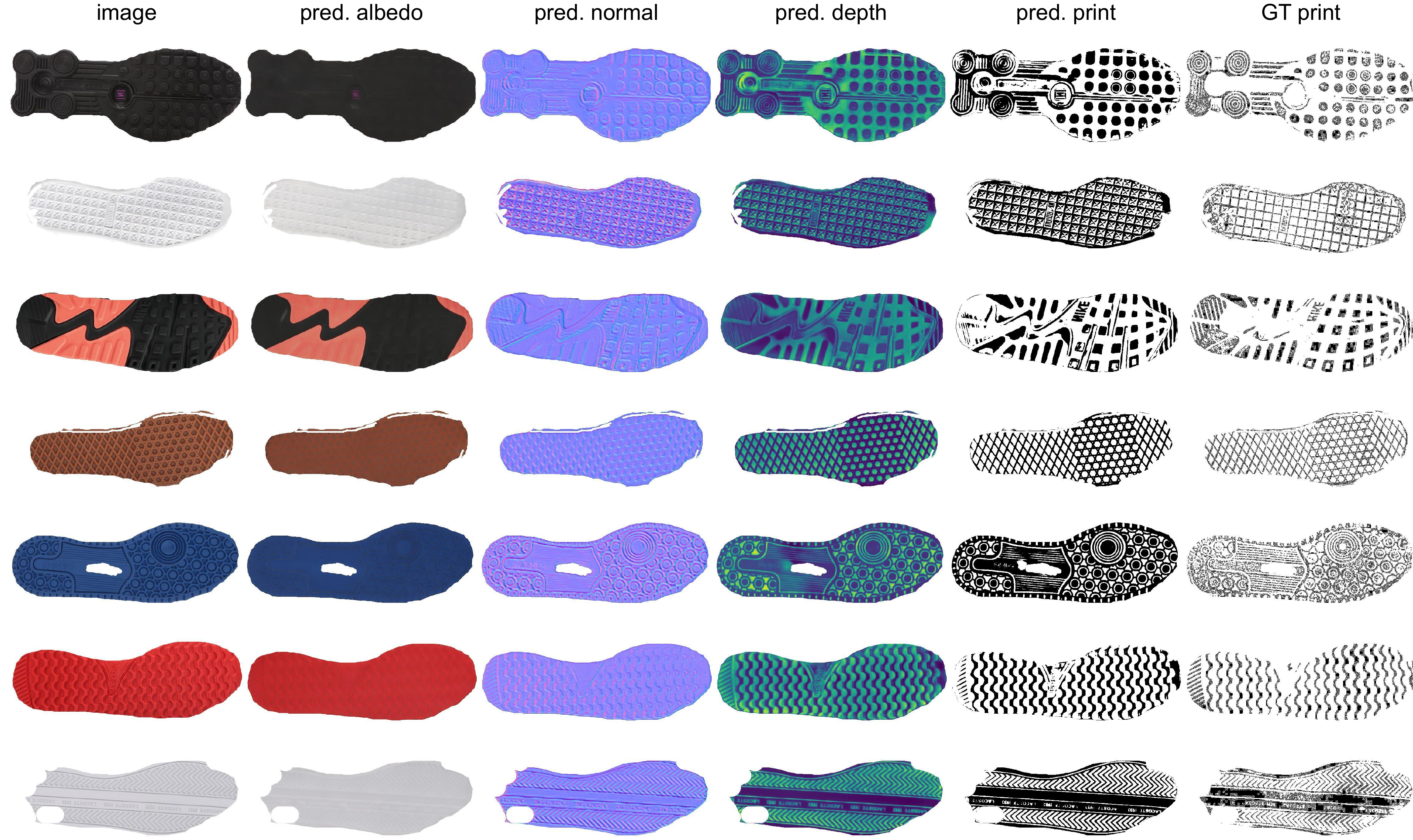}
\caption{\small
   Visualization of predicted shoeprints, as well as intrinsics, by our \emph{ShoeRinsics} on real-FID-val. Real-FID-val consists of images of real shoe-treads downloaded from online retailers and corresponding ground-truth shoeprints. 
   Visually, we can see our method works quite well w.r.t both shoeprint prediction and intrinsic decomposition (albedo, normal, and depth).
}
\label{fig:visual_for_real_shoe_ours}
\end{figure*}

\begin{figure*}[t]
\center
  \includegraphics[width=\linewidth]{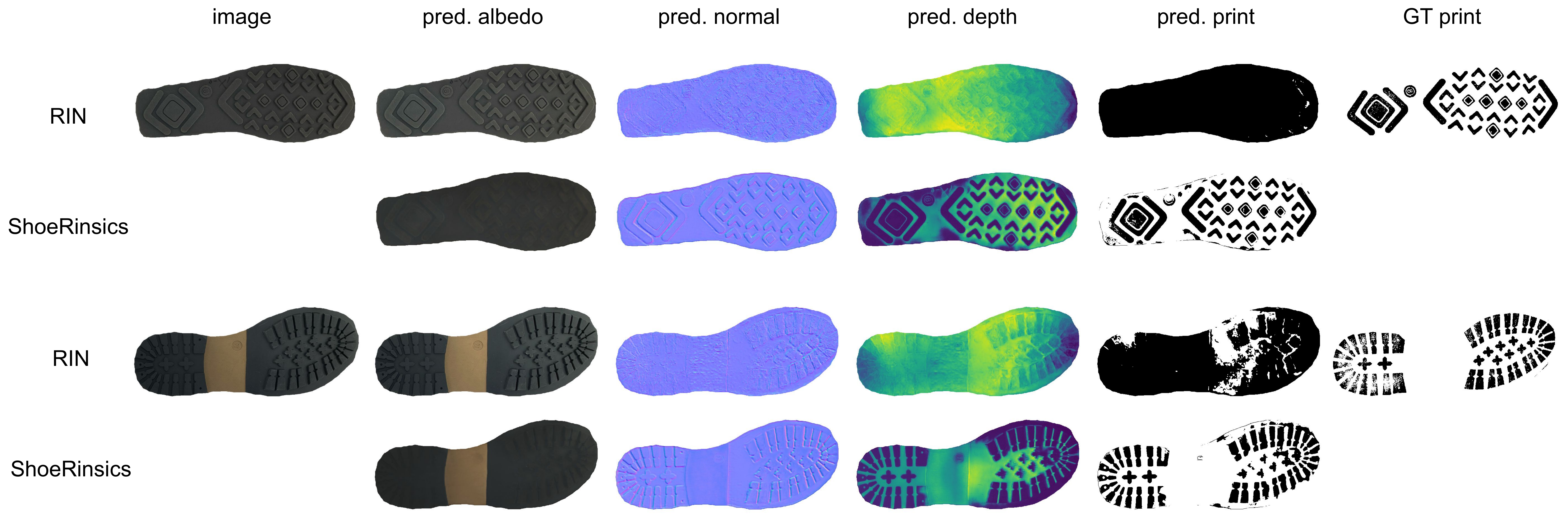}
\caption{\small
Qualitative comparison between RIN \cite{nips2017} and \emph{ShoeRinsics} on real-val. Along with input images and ground-truth shoeprint, we show albedo, normal, depth, and print prediction. Note that RIN does not directly produce depth predictions. We obtain them by integrating their normal predictions using the  well-established Frankot Chellappa algorithm \cite{frankot1988method}. As we can see, RIN produces poor quality albedo and normal predictions, presumably because it does not explicitly perform domain adaptation. 
The albedo predictions retain much of the shading information and the normal predictions are noisy. The substandard normal predictions in RIN lead to unsatisfactory depth and print predictions. Comparatively, \emph{ShoeRinsics} is able to produce more likely albedo, normal, and depth predictions and thus predict shoeprints which are much closer to the ground-truth.
}
\label{fig:RIN_visuals}
\end{figure*}

{\bf Threshold $t_{nc}$ for large non-contact regions.}
Although, our current best-matching shoeprint estimation is good enough for most cases, it may have issues for very large non-contact surfaces. Fig.~\ref{fig:t_nc_effect} illustrates how a large non-contact region can cause the local threshold $s d_l$ to be very high. This in turn can lead to incorrectly predicting some areas to leave a print (such as the logo in Fig.~\ref{fig:t_nc_effect}).   
We fix this by identifying the threshold $t_{nc}$ which gives the highest IoU between ground-truth shoeprint $S^*$  and  $S_{best} \; \text{AND} \; (\hat X^d_R < t_{nc})$ and update our shoeprint prediction $S_{best}$. 
We find it sufficient to determine $t_{nc}$ by sampling values in range $[0.1  p_{95}, p_{95}]$ at an interval of $0.01$ where $p_{95}$ is the $95^{th}$ percentile of sorted $\hat X^d_R$ values.

{\bf Threshold $t_c$ for large contact regions.}
A similar problem and solution apply for large contact surfaces as demonstrated in Fig.~\ref{fig:t_c_effect}. In such regions, the local threshold is very low and can result in ``holes'' in our predicted print. 
Our solution is to find threshold $t_{c}$ for which the IoU between ground-truth shoeprint $S^*$ and $S_{best}$ OR $(\hat X^d_R < t_c)$ is highest. 
We can find an adequate value for $t_c$ by sampling numbers from range $[p_{05}, 30 p_{05}]$ at intervals of $0.1$ where $p_{05}$ is the $5^{th}$ percentile of sorted $\hat X^d_R$ values.

{\bf Generic thresholds for print prediction.}
To determine shoeprint predictions from real images without ground-truth print, we set $s = 1$, $t_{nc} = p_{97}$, and $t_{c} = p_{03}$ where $p_x$ is the $x^{th}$ percentile of sorted $\hat X^d_R$ values.

\section{Qualitative Results of \emph{ShoeRinsics} on Real Shoe-Treads}
\label{sec:qualitaive_analysis}

In this section, we perform additional qualitative analysis of our \emph{ShoeRinsics}. We visualize our albedo, normal, depth, and print predictions (Fig.~\ref{fig:visual_for_real_shoe_ours}) as well as compare shoeprint predictions to that of RIN~\cite{nips2017} (Fig.~\ref{fig:RIN_visuals}).

\subsection{Visualization of Predictions on Real-FID-val}
\label{sec:results_real_test}

One of the datasets we collected, real-FID-val, consists of images of real shoe-treads downloaded from online retailers with corresponding ground-truth shoeprints.
We visualize our intrinsic predictions (albedo, normal, and depth) and compare print predictions to ground-truth shoeprints on the real-FID-val dataset in Figure~\ref{fig:visual_for_real_shoe_ours}. 
We see that the intrinsic predictions are visually pleasing and the predicted print closely resembles the ground-truth shoeprint.

\subsection{Comparison with RIN}
\label{sec:supp_rin}

RIN \cite{nips2017} learns from unlabeled real images using intrinsic image decomposition. It breaks down images into albedo, normal and light. We integrate their normal predictions to obtain a depth prediction using the well-established Frankot Chellappa algorithm \cite{frankot1988method}. Thresholding this depth prediction gives us the shoeprint prediction which we compare to the ground-truth shoeprint. 
In Figure~\ref{fig:RIN_visuals}, we compare the albedo, normal, depth, and shoeprint predictions of RIN on real-val with that of \emph{ShoeRinsics}. We find that RIN performs poorly on real shoes, presumably because it does not explicitly perform domain adaptation. Even though our focus is on the depth prediction, our albedo and normal predictions visually look better than the predictions made by RIN. Albedo predictions from RIN retain much of the shape information.
More importantly, noisy normal predictions from RIN integrate to give low quality depth predictions and thus unsatisfactory shoeprint predictions.

\begin{table}[H] 
\scriptsize
\caption{\small 
Comparison of Intersection over Union (IoU) values achieved by \emph{ShoeRinsics} and related work on each example in real-val. The best IoU per shoe example is written in bold and the second-best is underlined. 
We can see that \emph{ShoeRinsics w/ aug} is the clear winner while \emph{ShoeRinsics} is the second-best. 
}
\setlength\tabcolsep{0pt} 
\begin{tabular*}{\linewidth}{>{}p{1.3cm} @{\extracolsep{\fill}}*{6}{c}}
\toprule
Shoe ID & RIN & ADDA & UDAB & CyCADA & ShoeRinsics & ShoeRinsics \\ 
& \cite{nips2017} & \cite{adda} & \cite{udab} & \cite{cycada} &  & w/ aug \\ 
\midrule
0001-L	&	25.7	&	29.9	&	34.1	&	38.8	&	\underline{41.4}	&	\textbf{43.5}	\\
0001-R	&	24.2	&	22.8	&	29.0	&	33.0	&	\underline{34.3}	&	\textbf{36.8}	\\
0002-L	&	27.8	&	29.3	&	27.8	&	\textbf{35.5}	&	29.6	&	\underline{32.4}	\\
0002-R	&	26.4	&	28.8	&	28.5	&	\textbf{36.9}	&	33.7	&	\underline{35.0}	\\
0003-L	&	20.3	&	24.5	&	30.2	&	32.3	&	\underline{37.7}	&	\textbf{37.9}	\\
0003-R	&	21.4	&	22.7	&	25.8	&	28.7	&	\underline{33.0}	&	\textbf{33.7}	\\
0004-L	&	26.3	&	31.0	&	31.0	&	38.0	&	\underline{39.6}	&	\textbf{39.7}	\\
0004-R	&	23.3	&	28.5	&	30.1	&	32.6	&	\textbf{37.5}	&	\underline{35.9}	\\
0005-L	&	29.2	&	50.4	&	56.0	&	59.2	&	\textbf{60.5}	&	\underline{59.9}	\\
0005-R	&	27.1	&	54.3	&	60.0	&	56.6	&	\textbf{62.4}	&	\underline{60.6}	\\
0006-L	&	31.5	&	36.2	&	36.3	&	37.7	&	\underline{42.2}	&	\textbf{43.2}	\\
0006-R	&	30.0	&	34.7	&	36.0	&	\underline{38.5}	&	38.0	&	\textbf{40.4}	\\
0007-L	&	19.6	&	19.3	&	19.4	&	17.1	&	\underline{22.8}	&	\textbf{63.3}	\\
0007-R	&	23.1	&	24.0	&	24.0	&	22.6	&	\underline{32.5}	&	\textbf{42.4}	\\
0009-L	&	48.8	&	48.8	&	48.9	&	\textbf{58.4}	&	\underline{55.8}	&	49.7	\\
0009-R	&	47.3	&	50.5	&	48.9	&	\underline{55.6}	&	\textbf{56.1}	&	47.9	\\
0010-L	&	52.5	&	56.2	&	54.8	&	56.4	&	\textbf{66.6}	&	\underline{61.0}	\\
0010-R	&	46.8	&	49.4	&	46.5	&	\textbf{53.5}	&	\underline{53.2}	&	52.9	\\
0011-L	&	22.4	&	24.6	&	\underline{25.0}	&	\underline{25.0}	&	\underline{25.0}	&	\textbf{25.1}	\\
0011-R	&	25.7	&	\textbf{28.9}	&	28.2	&	\textbf{28.9}	&	28.5	&	\underline{28.8}	\\
0012-L	&	32.1	&	\underline{77.5}	&	68.2	&	75.8	&	76.2	&	\textbf{83.3}	\\
0012-R	&	30.5	&	72.3	&	69.6	&	\underline{72.5}	&	69.1	&	\textbf{76.9}	\\
0013-L	&	24.2	&	\textbf{35.7}	&	31.7	&	31.5	&	31.2	&	\underline{32.8}	\\
0013-R	&	27.5	&	\textbf{40.9}	&	36.6	&	38.2	&	37.9	&	\underline{39.0}	\\
0014-L	&	13.6	&	23.9	&	21.2	&	18.8	&	\underline{27.7}	&	\textbf{31.4}	\\
0014-R	&	15.3	&	29.4	&	29.0	&	22.1	&	\underline{32.4}	&	\textbf{37.8}	\\
0015-L	&	24.9	&	\textbf{36.8}	&	29.8	&	35.0	&	\underline{35.3}	&	34.4	\\
0015-R	&	27.4	&	45.5	&	41.1	&	41.8	&	\underline{53.1}	&	\textbf{53.2}	\\
0016-L	&	21.7	&	63.4	&	61.1	&	\underline{66.0}	&	63.9	&	\textbf{68.6}	\\
0016-R	&	21.4	&	61.3	&	60.5	&	\underline{65.3}	&	63.6	&	\textbf{67.8}	\\
0017-L	&	24.8	&	47.6	&	47.3	&	\underline{56.3}	&	55.3	&	\textbf{57.1}	\\
0017-R	&	26.4	&	47.8	&	54.2	&	\textbf{60.2}	&	\underline{59.2}	&	57.3	\\
0018-L	&	34.8	&	35.3	&	37.3	&	46.4	&	\textbf{51.8}	&	\underline{50.6}	\\
0018-R	&	37.9	&	38.6	&	38.0	&	48.9	&	\textbf{54.1}	&	\underline{52.8}	\\
0019-L	&	66.4	&	69.2	&	72.0	&	\underline{73.5}	&	71.4	&	\textbf{75.4}	\\
0019-R	&	66.1	&	68.7	&	73.2	&	\underline{75.2}	&	72.4	&	\textbf{76.0}	\\
\midrule
Average	&	30.4	&	41.4	&	41.4	&	44.8	&	\underline{46.8}	&	\textbf{49.0}	\\
\bottomrule
\end{tabular*}
\label{table:real-val-iou-details}
\end{table}

\section{Further Details of Quantitative Analysis}
\label{sec:quantitative_analysis_details}

We compare methods using our defined metric based on Intersection over Union (IoU). 
We analyse the IoU values for each of the shoe examples in real-val (Table~\ref{table:real-val-iou-details}) and real-FID-val (Table~\ref{table:real-FID-val-iou-details}) to further demonstrate that \emph{ShoeRinsics} outperforms the state-of-the-art domain adaptation and intrinsic image decomposition methods. We can see from the Tables that \emph{ShoeRinsics w/ aug} performs the best, followed by \emph{ShoeRinsics} in the second position.

\section{Synthetic Data Preparation}
\label{sec:syn_data}

To train our model, we need shoe-sole images with paired ground-truth albedo, depth, normal and light information. Publicly available datasets that include shoe objects (among other categories)~\cite{objectron2021} either do not focus on the shoe-sole and/or do not provide full decomposition into shape, albedo, and lighting. Thus, we introduce our own synthetic dataset, syn-train.

\begin{table}[H] 
\scriptsize
\caption{\small 
Comparison of Intersection over Union (IoU) values achieved by \emph{ShoeRinsics} and related work on each example in real-FID-val. The best IoU per shoe example is written in bold and the second-best is underlined. 
We can see that \emph{ShoeRinsics w/ aug} is the clear winner while \emph{ShoeRinsics} is the second-best. 
}
\setlength\tabcolsep{0pt} 

\begin{tabular*}{\linewidth}{>{}p{1.3cm} @{\extracolsep{\fill}}*{6}{c}}
\toprule
Shoe ID & RIN & ADDA & UDAB & CyCADA & ShoeRinsics & ShoeRinsics \\ 
& \cite{nips2017} & \cite{adda} & \cite{udab} & \cite{cycada} &  & w/ aug \\ 
\midrule
1	&	\textbf{33.0}	&	28.5	&	28.5	&	30.1	&	31.8	&	\underline{32.6}	\\
3	&	18.8	&	18.8	&	26.7	&	\textbf{35.9}	&	\underline{33.7}	&	33.4	\\
4	&	20.3	&	21.6	&	27.7	&	\underline{29.2}	&	27.9	&	\textbf{29.7}	\\
5	&	23.8	&	24.1	&	26.9	&	28.3	&	\textbf{29.9}	&	\underline{29.5}	\\
8	&	11.9	&	14.4	&	18.3	&	\textbf{21.4}	&	19.3	&	\underline{20.7}	\\
9	&	21.0	&	20.6	&	27.3	&	31.5	&	\underline{31.8}	&	\textbf{32.3}	\\
10	&	15.8	&	21.5	&	16.4	&	\textbf{23.8}	&	22.2	&	\underline{22.7}	\\
11	&	25.5	&	32.1	&	34.3	&	34.0	&	\underline{35.1}	&	\textbf{36.5}	\\
12	&	30.7	&	29.5	&	27.7	&	31.5	&	\textbf{32.5}	&	\underline{32.4}	\\
13	&	33.3	&	30.1	&	32.1	&	\underline{34.0}	&	33.2	&	\textbf{34.4}	\\
16	&	28.9	&	30.6	&	37.7	&	\textbf{52.4}	&	51.7	&	\underline{51.9}	\\
17	&	24.8	&	22.5	&	\underline{33.0}	&	\textbf{35.9}	&	29.9	&	29.6	\\
23	&	28.3	&	29.0	&	30.7	&	\textbf{32.8}	&	31.0	&	\underline{31.9}	\\
32	&	36.0	&	43.0	&	42.6	&	\textbf{47.0}	&	46.3	&	\underline{46.5}	\\
33	&	28.3	&	28.3	&	28.0	&	27.3	&	\underline{29.4}	&	\textbf{30.0}	\\
35	&	34.3	&	40.5	&	40.1	&	\underline{40.6}	&	\underline{40.6}	&	\textbf{41.2}	\\
45	&	31.2	&	30.8	&	33.3	&	32.8	&	\underline{36.9}	&	\textbf{37.2}	\\
47	&	\underline{24.8}	&	24.1	&	24.0	&	24.0	&	\textbf{24.9}	&	\textbf{24.9}	\\
53	&	11.7	&	11.8	&	11.7	&	12.1	&	\textbf{13.7}	&	\underline{13.1}	\\
54	&	22.0	&	22.0	&	22.0	&	22.1	&	\textbf{29.1}	&	\underline{29.0}	\\
55	&	30.6	&	28.7	&	29.4	&	30.1	&	\underline{31.6}	&	\textbf{31.8}	\\
56	&	19.0	&	\textbf{19.3}	&	\underline{19.2}	&	19.1	&	19.0	&	19.0	\\
62	&	33.6	&	33.9	&	36.3	&	\underline{36.8}	&	\textbf{38.1}	&	\textbf{38.1}	\\
72	&	\underline{28.2}	&	28.1	&	\underline{28.2}	&	\textbf{29.0}	&	\underline{28.2}	&	\underline{28.2}	\\
74	&	32.8	&	34.1	&	35.5	&	34.7	&	\underline{36.3}	&	\textbf{36.8}	\\
82	&	44.2	&	36.8	&	41.3	&	\underline{45.5}	&	45.3	&	\textbf{45.7}	\\
1040	&	42.0	&	37.8	&	40.3	&	45.4	&	\underline{46.0}	&	\textbf{47.1}	\\
1041	&	27.6	&	36.7	&	35.5	&	35.3	&	\underline{38.0}	&	\textbf{38.1}	\\
1044	&	19.5	&	20.7	&	20.6	&	21.5	&	\underline{23.9}	&	\textbf{24.3}	\\
1047	&	29.1	&	29.2	&	30.2	&	\underline{31.2}	&	\textbf{31.4}	&	\textbf{31.4}	\\
1048	&	23.5	&	27.7	&	\underline{28.5}	&	28.1	&	28.0	&	\textbf{28.6}	\\
1049	&	26.7	&	21.6	&	25.3	&	27.7	&	\underline{27.8}	&	\textbf{28.2}	\\
1050	&	\underline{26.4}	&	25.4	&	\underline{26.4}	&	26.1	&	\textbf{26.5}	&	\underline{26.4}	\\
1058	&	24.2	&	36.5	&	\textbf{38.8}	&	35.2	&	36.7	&	\underline{38.0}	\\
1062	&	19.0	&	21.0	&	23.5	&	25.6	&	\textbf{29.1}	&	\underline{28.6}	\\
1064	&	18.0	&	20.7	&	23.4	&	21.6	&	\underline{24.4}	&	\textbf{24.6}	\\
1071	&	11.7	&	16.5	&	\underline{19.2}	&	\textbf{19.8}	&	18.3	&	18.3	\\
1076	&	24.3	&	30.0	&	30.1	&	\underline{33.6}	&	\textbf{34.5}	&	\underline{33.6}	\\
1079	&	26.4	&	28.4	&	29.4	&	29.3	&	\underline{31.3}	&	\textbf{31.4}	\\
1088	&	29.2	&	27.4	&	29.9	&	33.5	&	\underline{34.0}	&	\textbf{34.2}	\\
1095	&	26.8	&	29.7	&	28.7	&	37.7	&	\underline{38.1}	&	\textbf{41.4}	\\
\midrule
Average	&	26.0	&	27.2	&	29.0	&	31.1	&	\underline{31.6}	&	\textbf{32.0}	\\
\bottomrule
\end{tabular*}
\label{table:real-FID-val-iou-details}
\end{table}

For this purpose, we synthesize  depth maps, albedo maps, and lighting environments. 
We observe that commercial shoe tread photographs are taken under very diffuse lighting conditions where the primary variations in surface brightness are driven by global illumination effects rather than surface normal orientation (e.g., grooves appear darker). This necessitates the use of a physically-based rendering engine~\cite{Mitsuba} rather than simple local shading models.
We discuss details of depth map generation in Section~\ref{sec:syn_data_depth} and visualize the different light environment maps in Section~\ref{sec:syn_data_light}

\begin{figure}[H]
\center
  \includegraphics[width=\linewidth,trim={0 0.5cm 36cm 0},clip]{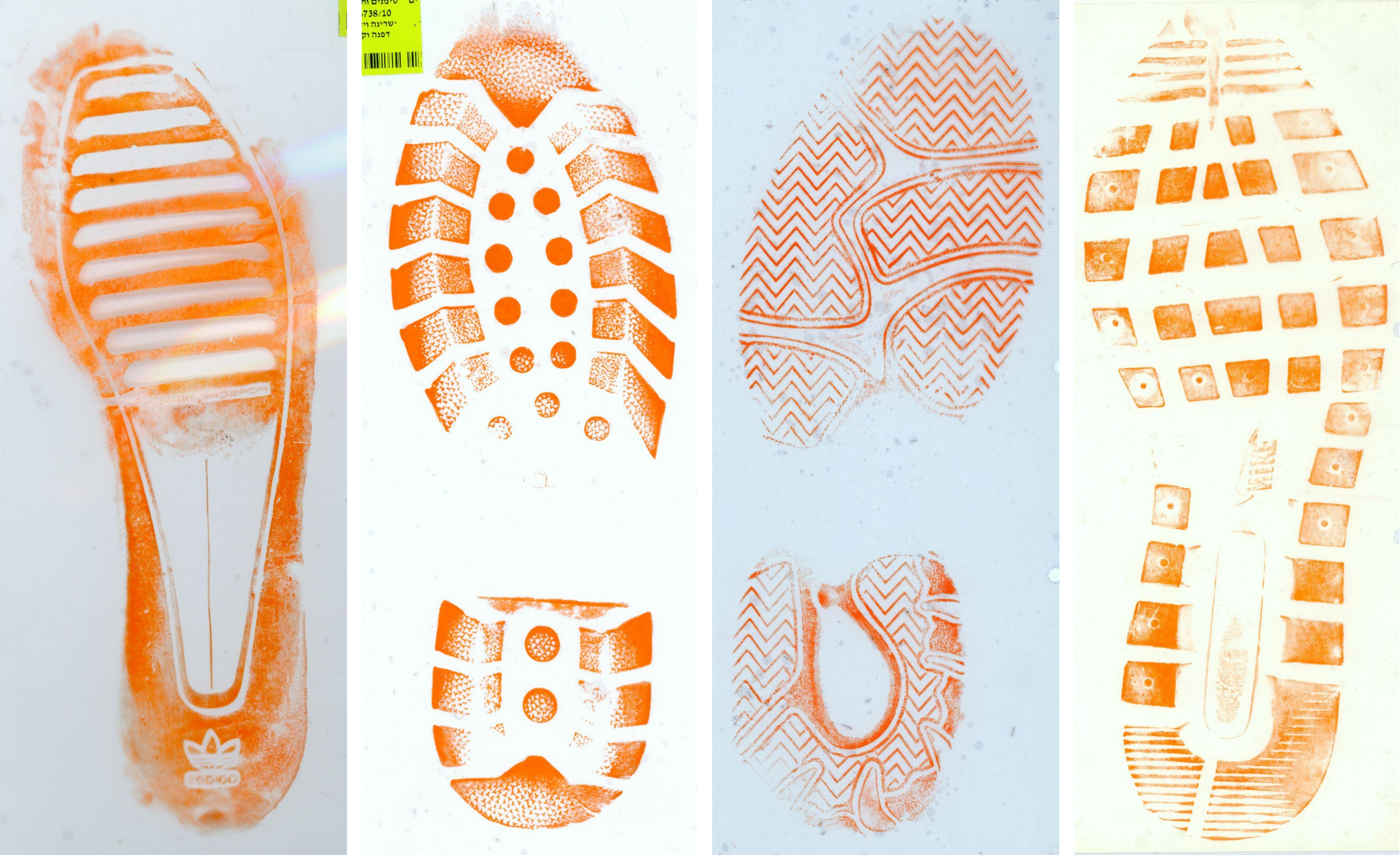}
\caption{\small
Examples from an existing shoeprint dataset~\cite{isreal-shoe}. We use these prints as a starting point for synthetic depth map generation. Note that even though these are shoeprints collected under a controlled lab environment, the images are still quite noisy. This necessitates some preprocessing before these can be used for synthetic depth map generation.
}
\label{fig:israeli_shoeprint}
\end{figure}

\subsection{Depth Map Generation}
\label{sec:syn_data_depth}

\begin{figure*}[t]
\center
  \includegraphics[width=\linewidth,trim={0 6.7cm 27.9cm 0},clip]{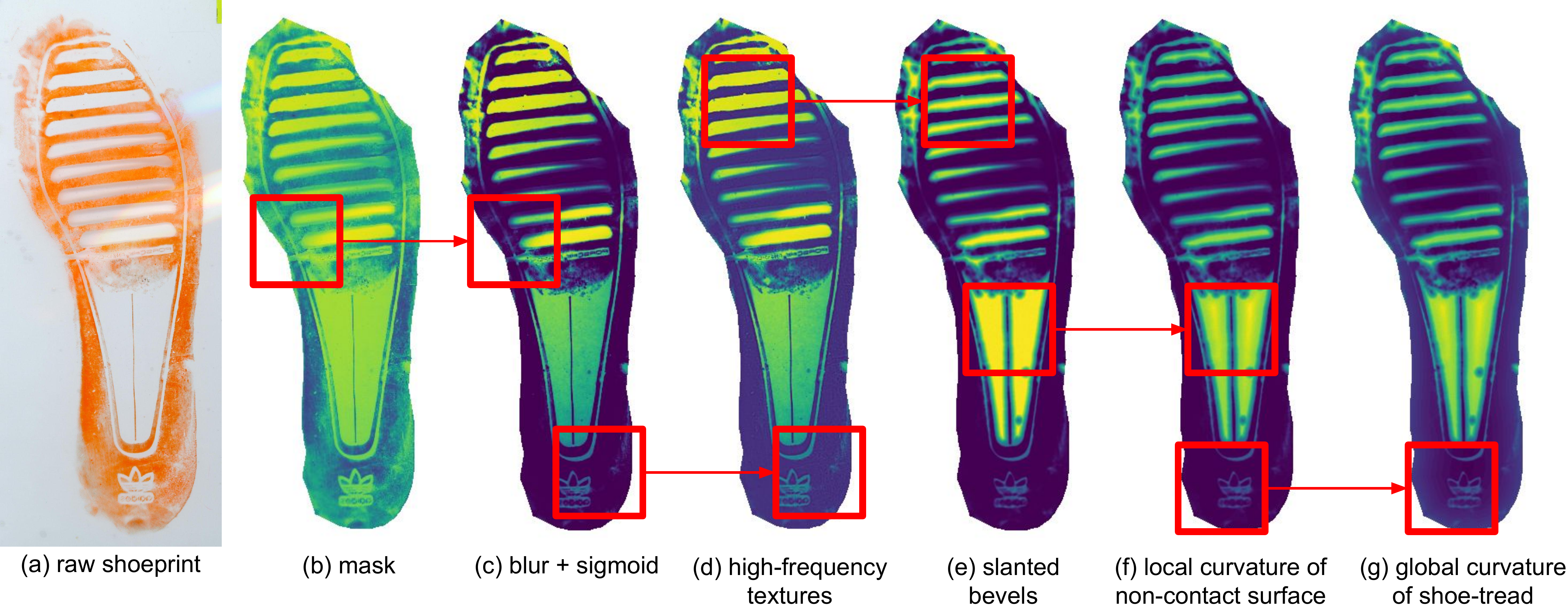}
\caption{\small
We illustrate major steps in depth map generation with an example. We first filter noise in a shoeprint image (a) by masking out the background (b) and applying a Gaussian blur followed by a sigmoid function (c) on the shoeprint image. Then, we add in some moderated amount high frequency details from the shoeprint image as textures (d). To make our depth maps more realistic, we optionally add slanted bevels (e), local curvatures for non-contact surfaces (f), and a global curvature along the edge of the shoe-tread (g). 
}
\label{fig:syn_depth_map}
\end{figure*}

We use an existing shoeprint dataset~\cite{isreal-shoe} collected in a controlled lab environment. Sample shoeprints are displayed in Fig.~\ref{fig:israeli_shoeprint}. We convert the 2D shoeprints to 3D depth maps by adding fake depth values to each point on the print. We generate 10-15 different depth maps from each of the 387 shoeprints available in~\cite{isreal-shoe}. Fig.~\ref{fig:syn_depth_map} highlights major steps in depth map generation from shoeprint images. Details of depth map generation is provided below. 

{\bf Removing noise.}
Raw shoeprint data is noisy (as shown in Fig~\ref{fig:israeli_shoeprint}).  
We employ two tricks to filter noise. First, we compute a mask for the shoeprint to remove notes and dirt in the background from consideration. Given that the shoeprints are orange colored and the background does not contain any of that color, we determine the mask using the concave hull of the orange colored regions. 
Second,  we filter noise by applying a Gaussian blur followed by a sigmoid function on the gray-scale shoeprint. 

{\bf Adding a realistic touch.}
At this stage, our depth map mainly consists of the two extreme values representing contact and non-contact surfaces. To incorporate some texture, we add a moderated amount of high frequency details (obtained from subtracting the blurred depth from the original gray-scale shoeprint).  
Next, we optionally add slanted bevels to our depth map to make the tread blocks look more natural. 
We further add a local curvature to the non-contact surfaces to give them some dimension. Essentially, we square the euclidean distance transform of the depth image and add the smoothed out result to our depth map. 
Finally, we also add a global curvature along the edge of the shoe-tread to attain the natural upward curvature that is common in many shoes.

\begin{figure*}[!th]
\hspace{19mm} light \hspace{18mm}  rendered shoe \hspace{28mm} light \hspace{18mm}  rendered shoe
\vspace{-3mm}
\center
\includegraphics[height=19.5cm,trim={0 0 57.9cm 0},clip]{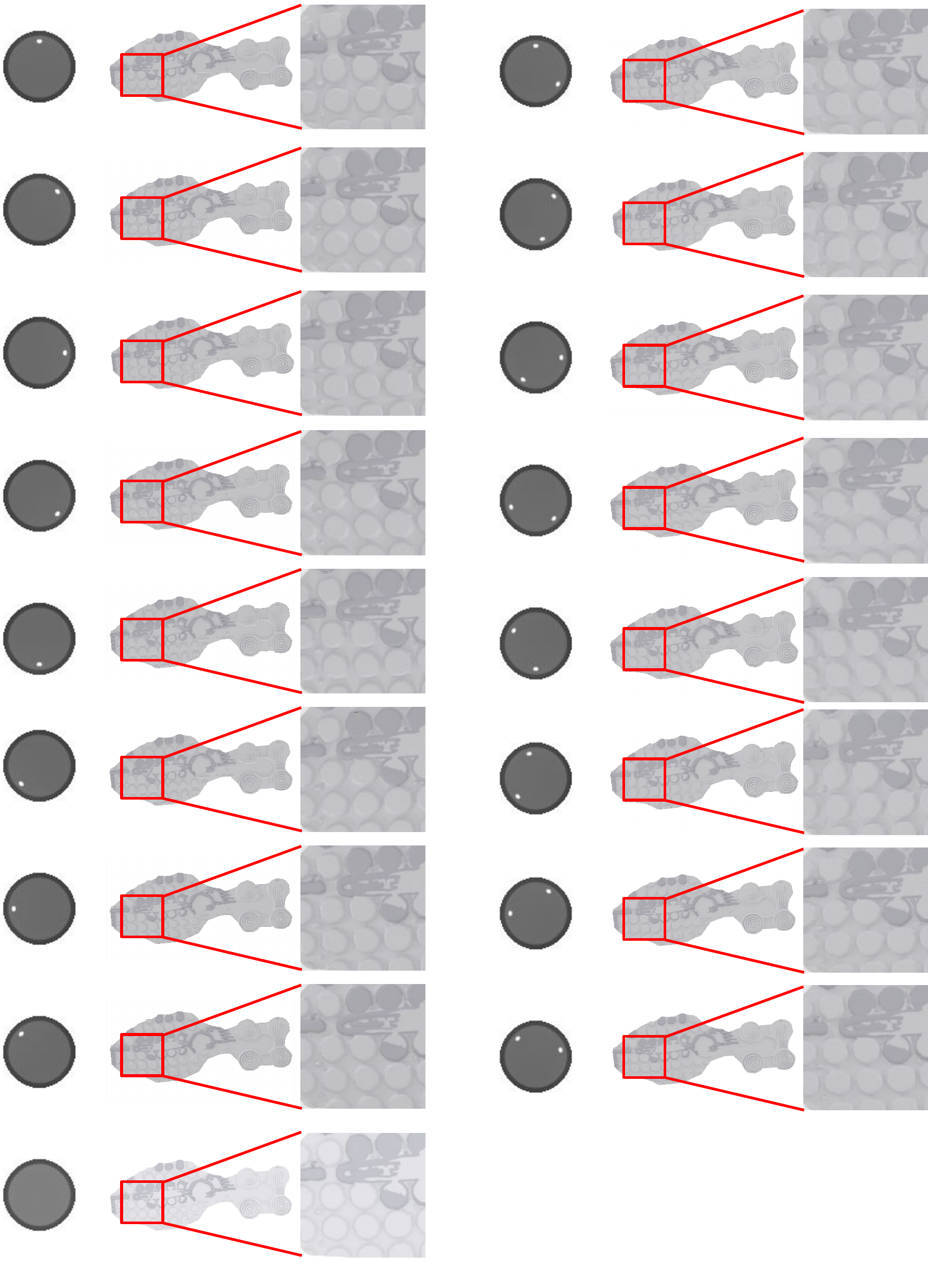}
\caption{\small
Visualization of the 17 different light types in our synthetic dataset (syn-train). We show a shiny sphere representing the light in the environment and a shoe rendered under that light condition. 
Our light environments consist of diffuse white light in addition to 0 to 2 light bulbs for directional light.
Different light sources produce different shadows on the shoe-tread blocks. Please refer to the attached video for a better visualization.
}
\label{fig:syn_shoe_light_visual}
\end{figure*}

\subsection{Visualization of Light Environments}
\label{sec:syn_data_light}

We display the different light environments in our synthetic dataset (syn-train) in Figure~\ref{fig:syn_shoe_light_visual} and also provide a video to better visualize the lighting effects. We have a total of 17 different light environments in our dataset. One consists of diffuse white light. Eight light configurations consist of a white light bulb providing directional light (from 8 different directions) in addition to the diffuse white light. 
The other eight light configurations consists of two white light bulbs at a $120^\circ$ angle to each other in addition to the diffuse white light.  
For each light configuration, we visualize a shiny sphere in place of the shoe demonstrating the light placements. Additionally, we render an example synthetic shoe under all the different light conditions to show the effect of each of them. We see different light environments create different shadows on on the shoe-tread blocks.   

\section{Real Data Preparation for Evaluation}
\label{sec:supp_gt_print}

To quantitatively evaluate and compare methods, create real-val which consists of paired shoe-tread images and ground-truth prints  for real shoes.

{\bf Photographing shoe-treads.}
Real-val contains new-athlectic, used-athletic, and new-formal shoes. The new shoes are collected from thrift stores which often sell new or very lightly used shoes. The used shoes are worn-out athletic shoes donated by volunteers. We first clean all the shoe-treads with soap and water and let them dry. 
Next, we photograph the shoe-treads in a brightly lit environment similar to that of a professional photography setting. We put together 5 square light panels to create a light box and place the shoe on a holder inside the light box. We also illuminate the shoes using a ring light on top. 

{\bf Preparing ground-truth shoeprint.}
After photographing the shoes we proceed to collecting their prints. We use a process called \emph{block printing technique} 
which is widely used in forensics to collect lab shoeprint impressions \cite{bodziak2017footwear}. With the shoe resting on the holder, we paint the shoe-tread with a thin layer of relief ink using a roller. Forgoing the roller and simply using a paint brush would cause ink blobs to get stuck in the nooks and crannies of the shoe-tread leading to blotchy prints. While the ink is still wet, we quickly press a slightly absorbent white paper onto the shoe-tread using a roller. The use of the roller distributes pressure throughout the paper and thus produces more uniform prints. 
We collect 2-3 sets of prints for each shoe, each time painting the shoe with a new layer of ink. Notice how these individual prints are not identical and  contain areas of uneven coverage. To get a smoother result, we align all the prints to the shoe-tread image (using thin-plate spline~\cite{duchon1977splines} and point correspondences between the shoe-tread image and the collected prints) and average them. The average is a more complete and evenly colored print. Finally, the average print is thresholded to get our binary ground-truth shoeprint.

\section{Further Implementation Details}
\label{sec:supp_implementation}

{\bf Decomposer $\mathcal F$.}
Our decomposer consists of a classic encoder-decoder structure with skip connections. We use separate decoders for albedo, normal, depth, and light predictions. All of the encoded input is passed to each of these decoders. The light decoder consists of residual blocks followed by a final convolution layer which outputs 17 numbers representing the probability of predicting the 17 light types in our synthetic  training dataset. We use the output of the second last layer of the albedo, normal, and depth decoder as the corresponding features. For light features, we use the 17 light probabilities. 

{\bf Renderer $\mathcal R$.}
The renderer has a mirrored structure as the decomposer. It has separate encoders for albedo, depth, normal, and light. The light encoder takes in a one-hot array representing the light configuration. The encoded information from each of the encoders is concatenated and passed to the decoder which predicts the synthetic or real shoe-tread image. 

{\bf Connecting the decomposer and renderer.}
When passing decomposer outputs to the renderer in our main pipeline, we ensure that the decomposer outputs look similar to the synthetic albedo, depth, normal, and light used to train the renderer. We set the background (i.e., parts outside the shoe-tread) pixel values to 1 in the albedo, depth, and normal predictions. We also use the Hard-Gumble trick to represent the light predictions as one-hot vectors instead of fractional probabilities for the renderer. This ensures that a path exists for gradient back-propagation through the light decoder while providing a one-hot representation for the light probabilities. 

{\bf Image translators $\mathcal G_{S\rightarrow R}$ and $\mathcal G_{R\rightarrow S}$.}
We use a ResNet backbone for the image translators. The two generators have the exact same structure. They consist of 2 convolution layers with stride 2, followed by 9 residual blocks, and finally 2 convolution layers coupled with nearest 2D upsampling layers with a scale factor of 2. The convolution layers and residual blocks in the generators are interspersed with batch normalization and the leakyReLU activation function.

{\bf Image discriminators $\mathcal D_S $ and $\mathcal D_R$.}
The discriminators used to learn image translation are PatchGANs and consist of 4 convolution layers with stride 2 followed by 2 convolution layers with stride 1. Similar to the image translators, the discriminators also have batch normalization and the leakyReLU activation function interspersed among the convolution layers and residual blocks.

{\bf Feature discriminator $\mathcal D_{feat}$.}
The discriminator for feature alignment takes in the concatenation of the albedo, normal, and depth features as one input, and the light features as a second input. These features are processed in two separate branches and the results are concatenated in the final output. Each of these branches consist of 3 convolution layers. The branch for the albedo, normal, and depth features uses a  kernel size of 3 (to encode some context), while the branch for light features use a kernel size of 1. 


\begin{figure}[H]
\center
  \includegraphics[width=\linewidth,trim={0 15.8cm 66cm 0},clip]{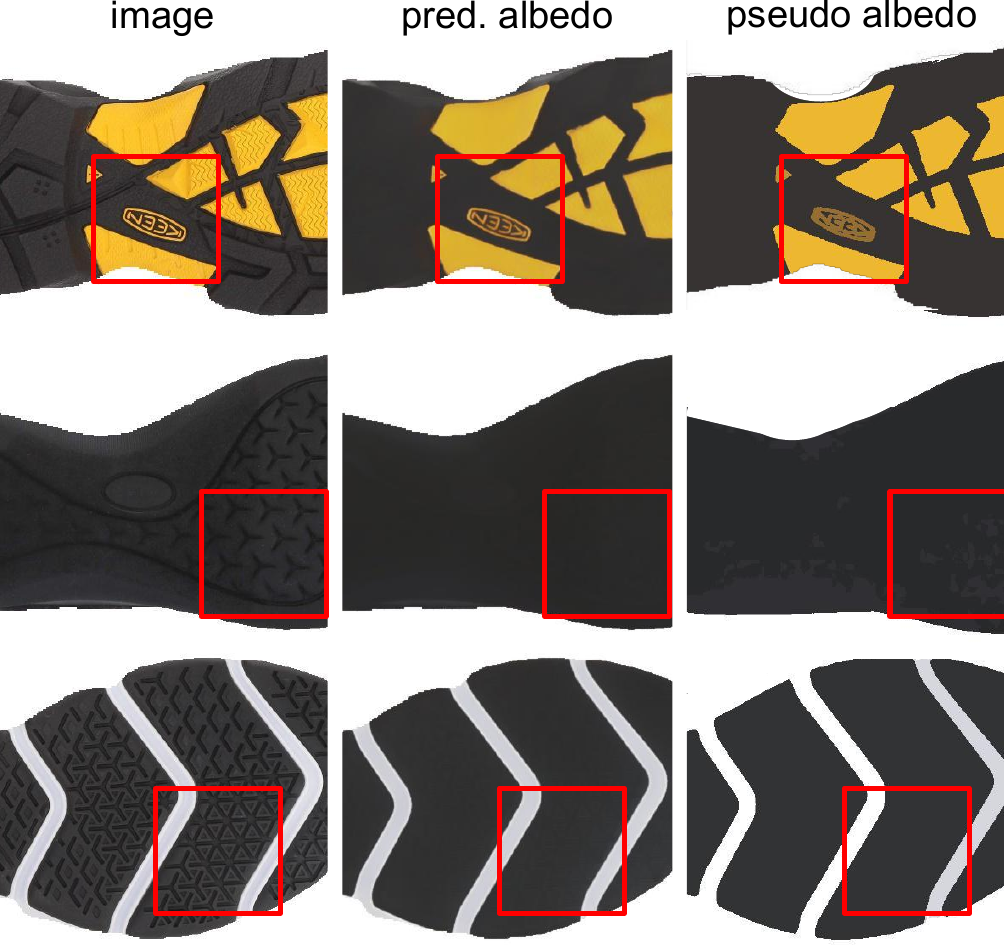}
\caption{\small
Visualization of difference between predicted albedo and pseudo albedo. Given that the albedo for shoe-treads consists mostly of piece-wise constant segments, we use the mean shift clustering algorithm~\cite{fukunaga1975estimation} to determine pseudo albedo. \emph{ShoeRinsics} learns to predict albedo for real shoe-treads using the psuedo albedo as ground-truth. 
We do not use pseudo albedo directly instead of the albedo prediction because it is not perfect ground truth and contains deviating segment boundaries (row 1), over-segmentation (row 2), and incorrect albedo labels for segments (row 3).
Our \emph{ShoeRinsics} learns to fix these errors.
}
\label{fig:pseudo_albedo_vs_predicted_albedo}
\end{figure}

\section{Discussion on the Pseudo Albedo}
\label{sec:supp_pseudo_albedo}

We provide pseudo supervision on the albedo prediction of real images. Fig.~\ref{fig:pseudo_albedo_vs_predicted_albedo} shows examples of pseudo albedo and the albedo predictions made by \emph{ShoeRinsics} on real shoe-tread images. The following is a discussion on pseudo albedo generation and how pseudo albedo differs from predicted albedo.

\subsection{Pseudo Albedo Generation}
{\bf Creating pseudo albedo segments.}
We first group the pixels in the real image using the mean-shift algorithm \cite{fukunaga1975estimation}. 
To generate the pseudo albedo labels, we work with the LAB color space since it is easier to distinguish hue ($A$ and $B$) from brightness ($L$) in this color space. Additionally, to ensure that shading does not interfere with pixel grouping, we scale the $L$ channel by a factor of $0.15$. Note that ignoring $L$ altogether would make it difficult to distinguish between black and white.
It turns out that we do not need to work on full resolution images for pixel grouping. So, we first downsize our real-shoe images 
to $67 \times 150$ for faster computation.  
After running mean-shift on the resulting shoe-tread pixels, we get 
an initial segmentation of the pixels in the real image. We define the color of each segment as the average color across that segment.


{\bf Refining pseudo albedo segments.}
The initial segmentation is very grainy as expected. Thus, we proceed to iteratively refine the segments for a maximum of $10$ iterations. For each iteration we merge `nearby' segments and update the color of the segments to reflect the segment updates. 
To merge segments, we find segments which are small in size and close to another segment both physically (share segment boundary) and numerically (have similar segment color). After merging segments, we update the color of the resulting segment as the average color across all the pixels in the new segment.
We break the iterative refinement loop when we reach an iteration where the segmentation does not receive any updates or when the maximum iteration count ($10$) has been reached. Since this is a time-consuming process, we predetermine the pseudo albedo for all real shoes and save them to be used directly during training. 





\subsection{Comparing Pseudo Albedo to Predicted Albedo}

It may seem counter-intuitive to learn to predict albedo when we can simply determine the corresponding ground-truth pseudo albedo. However, as we can see in Fig.~\ref{fig:pseudo_albedo_vs_predicted_albedo}, pseudo albedo is only approximate and can contain deviating segment boundaries (row 1), over-segmentation (row 2), and incorrect albedo labels for segments (row 3). \emph{ShoeRinsics} learns to fix these errors when trained using pseudo albedo as ground-truth.


\end{multicols}

\end{document}